\documentclass[journal]{IEEEtran}

\ifCLASSOPTIONcompsoc
  \usepackage[nocompress]{cite}
\else
  \usepackage{cite}
\fi

\usepackage{amsfonts,dsfont}
\usepackage{balance}
\usepackage{booktabs}
\usepackage{amsmath}
\usepackage{tabularx}
\usepackage{subfigure}
\usepackage{graphicx}
\usepackage{multirow}
\usepackage{rotating}
\usepackage{algorithm}
\usepackage{algorithmic}
\usepackage{url}
\usepackage{amssymb}
\usepackage{amsthm}
\usepackage{enumerate}
\usepackage{epsfig}
\usepackage{epstopdf}
\usepackage{tikz}
\usepackage{adjustbox}
\usepackage{color}% This package is for using color package.
\newtheorem{theorem}{Theorem}
\newtheorem{lemma}{Lemma}

\begin{document}
%
% paper title
% Titles are generally capitalized except for words such as a, an, and, as,
% at, but, by, for, in, nor, of, on, or, the, to and up, which are usually
% not capitalized unless they are the first or last word of the title.
% Linebreaks \\ can be used within to get better formatting as desired.
% Do not put math or special symbols in the title.
\title{Gray Learning from Non-IID Data with Out-of-distribution Samples}

\author{Zhilin Zhao,~% ~\IEEEmembership{Student Member,~IEEE,}\\
        Longbing Cao,~\IEEEmembership{Senior Member,~IEEE,}
        and~Chang-Dong Wang,~\IEEEmembership{Senior Member,~IEEE}% <-this % stops a space
\thanks{This work was supported in part by the Australian Research Council Discovery under Grant DP190101079 and in part by the Future Fellowship under Grant FT190100734.}
\IEEEcompsocitemizethanks{\IEEEcompsocthanksitem Zhilin Zhao and Longbing Cao are with the Data Science Lab, School of Computing and DataX Research Centre, Macquarie University, Sydney, NSW 2109, Australia. %\protect\\
%Zhilin Zhao and Longbing Cao are with the Advanced Analytics Institute, School of Compute Science, University of Technology Sydney, NSW 2007, Australia.\protect\\
% note need leading \protect in front of \\ to get a newline within \thanks as
% \\ is fragile and will error, could use \hfil\break instead.
E-mail: zhaozhl7@hotmail.com, longbing.cao@gmail.com
\IEEEcompsocthanksitem Chang-Dong Wang is with the School of Computer Science and Engineering, Sun Yat-sen University, Guagnzhou, China, Guangdong Province Key Laboratory of Computational Science, Guangzhou, China,
and Key Laboratory of Machine Intelligence and Advanced Computing, Ministry of Education, China. E-mail: changdongwang@hotmail.com
}% <-this % stops an unwanted space
}

% The paper headers
\markboth{}%
{\MakeLowercase{\textit{Zhao et al.}}: GL}

% make the title area
\maketitle

% As a general rule, do not put math, special symbols or citations
% in the abstract or keywords.
\begin{abstract}
The integrity of training data, even when annotated by experts, is far from guaranteed, especially for non-IID datasets comprising both in- and out-of-distribution samples. In an ideal scenario, the majority of samples would be in-distribution, while samples that deviate semantically would be identified as out-of-distribution and excluded during the annotation process. However, experts may erroneously classify these out-of-distribution samples as in-distribution, assigning them labels that are inherently unreliable. This mixture of unreliable labels and varied data types makes the task of learning robust neural networks notably challenging. We observe that both in- and out-of-distribution samples can almost invariably be ruled out from belonging to certain classes, aside from those corresponding to unreliable ground-truth labels. This opens the possibility of utilizing reliable complementary labels that indicate the classes to which a sample does not belong. Guided by this insight, we introduce a novel approach, termed \textit{Gray Learning} (GL), which leverages both ground-truth and complementary labels. Crucially, GL adaptively adjusts the loss weights for these two label types based on prediction confidence levels. By grounding our approach in statistical learning theory, we derive bounds for the generalization error, demonstrating that GL achieves tight constraints even in non-IID settings. Extensive experimental evaluations reveal that our method significantly outperforms alternative approaches grounded in robust statistics.
\end{abstract}

% Note that keywords are not normally used for peerreview papers.
\begin{IEEEkeywords}
Non-IID Data, Out-of-distribution Data, Gray Learning, Complementary Label, Generalization.
\end{IEEEkeywords}

% For peer review papers, you can put extra information on the cover
% page as needed:
% \ifCLASSOPTIONpeerreview
% \begin{center} \bfseries EDICS Category: 3-BBND \end{center}
% \fi
%
% For peerreview papers, this IEEEtran command inserts a page break and
% creates the second title. It will be ignored for other modes.
\IEEEpeerreviewmaketitle

\section{Introduction}
\label{sec:introduction}
\IEEEPARstart{D}{eep} neural networks trained on \textit{in-distribution} data demonstrate powerful generalization capabilities when tested on samples from the same distribution~\cite{NN:17,NN:19}. The training of such networks often necessitates large volumes of labeled data. To accumulate this data, original samples must be collected from various sources and subsequently annotated by experts~\cite{CS:18,PAC:17}. However, this raw data pool is not guaranteed to be clean. It may include \textit{out-of-distribution}~\cite{FIG:22, UE:22} samples with semantic shifts originating from other distributions. These out-of-distribution samples do not belong to any classes within the in-distribution dataset and should, therefore, be discarded during the annotation process~\cite{OOD:19}. Despite this, due to limitations in expert knowledge or inadvertent errors, these out-of-distribution samples can be misclassified as in-distribution and erroneously labeled~\cite{CS:14}. As a result, the acquired training dataset becomes non-independent and identically distributed (non-IID), incorporating both in-distribution and out-of-distribution samples. Such non-IID data can distort the classification learning for in-distribution samples and inevitably impair the generalization capabilities of the trained network~\cite{DN:17, US:19, GA:12}. Therefore, deriving a robust network from such contaminated non-IID data is of paramount importance.

% challenges
The primary challenge in training a robust network from non-IID data containing both in-distribution and out-of-distribution samples lies in leveraging the reliable information embedded within the unclean dataset. While the class labels for in-distribution samples are generally trustworthy, those for out-of-distribution samples are not. However, a network initialized randomly lacks the capacity to discern which samples are in-distribution and which are out-of-distribution. This presents a dilemma: Directly learning to classify by mapping inputs to their ground-truth labels would mislead the learning process because the out-of-distribution samples do not belong to the classes corresponding to their ground-truth labels. Nonetheless, we find that the complementary labels\footnote{Contrary to the definition of ground-truth labels that refer to the class a given input belongs to, complementary labels refer to the classes a given input does not belong to.}~\cite{CL:17,CL:22} for a given sample, be it in-distribution or out-of-distribution, are consistently reliable. Specifically, an in-distribution sample does not belong to any classes other than the one corresponding to its ground-truth label, and an out-of-distribution sample does not belong to any class. Therefore, the reliable information in this unclean non-IID dataset resides in the complementary labels. Based on this insight, \textit{classification can be indirectly learned by training the network to reject mapping a sample to its corresponding complementary labels.}

%methods
Based on the above discussion, we propose a novel method, referred to as \textit{gray learning} (GL), specifically designed for learning robust networks from non-IID data that includes both in-distribution and out-of-distribution samples. The core principle of GL is to learn from both ground-truth and complementary labels while adaptively adjusting the weights of the losses for these two types of labels based on prediction confidence. To elaborate, GL focuses on leveraging the ground-truth label for samples likely to be in-distribution, and complementary labels for those likely to be out-of-distribution. GL employs Maximum over Softmax Probabilities (MSP)~\cite{BL:17} as a mechanism to calculate prediction confidence, which serves as a proxy for the likelihood that a given sample is in-distribution. Higher probabilities suggest that a sample is more likely to be in-distribution, whereas lower probabilities indicate the opposite. For each training sample, GL computes two distinct losses: one for its ground-truth label and another for its complementary labels. These loss weights are then adaptively adjusted based on the prediction confidence. During the early training stages, prediction confidences for all samples are generally low due to the network's limited classification knowledge. As a result, the network primarily learns from the complementary labels for both in-distribution and out-of-distribution samples. However, as the training progresses, prediction confidence for in-distribution samples improves, widening the confidence gap between in- and out-of-distribution samples. This allows the network to increasingly rely on ground-truth labels for high-confidence samples that are more likely to be in-distribution, thereby further refining its ability to discriminate between in- and out-of-distribution samples based on their confidence levels.

The major contributions of this study are as follows:
\begin{enumerate}
\item A groundbreaking \textit{gray learning} (GL) framework is proposed to effectively learn from non-IID data sets, which comprise a mix of in-distribution and out-of-distribution samples.
\item The generalization error bound is derived to elucidate the impact of out-of-distribution samples on model performance and to provide assurances of convergence.
\item Comprehensive experiments demonstrate that the proposed approach significantly surpasses existing methods grounded in robust statistics.
\end{enumerate}

The remainder of the paper is organized in the following manner: Section~\ref{sec:rw} offers an overview of related work. Section~\ref{sec:tpm} details the methodology of the proposed GL framework. Theoretical underpinnings and empirical results are presented in Sections~\ref{sec:the} and~\ref{sec:exp}, respectively. Finally, Section~\ref{sec:con} provides concluding remarks.

\section{Related Work} \label{sec:rw}
Non-IID data presents various non-IIDness and non-IID settings, becoming a critical challenge in both shallow and deep learning \cite{Cao14,Pentina15,JianCLG18,PangCC21,Luo21,ZhuCY22}. In this work, we explore a novel non-IID scenario characterized by the presence of both in-distribution and out-of-distribution samples. As there are no existing studies that specifically address this unique scenario, we draw insights from various related research fields that also grapple with non-IID data complexities. These include but are not limited to, out-of-distribution detection~\cite{OOD:19}, outlier detection~\cite{OD:20}, domain generalization~\cite{DBLP:journals/pami/ZhouLQXL23}, learning with noisy labels~\cite{NL:19}.

\subsection{Out-of-distribution Detection}
Out-of-distribution detection aims to distinguish between in-distribution and out-of-distribution samples during the
test phase~\cite{ODIN:18} of networks, and various methods have been proposed for this purpose. Maximum over Softmax Probabilities (MSP)~\cite{BL:17} uses a threshold-based detector that gauges sample confidence through maximum softmax outputs. Prior Networks (PN)~\cite{PN:18} enhances this discrimination by incorporating real-world out-of-distribution samples during training to widen the confidence gap between the two types of samples. Confidence-Calibrated Classifier (CCC)~\cite{GO:18} employs a generative adversarial network to establish boundary points for in-distribution samples and treats these boundaries as indicative of out-of-distribution, thereby encouraging the network to assign lower confidence to such samples. IsoMax~\cite{UE:22} refines MSP by resolving issues related to anisotropy and low entropy. Learning from Cross-class Vicinity Distribution (LCVD)~\cite{LCVD:23} aims to diminish confidence in out-of-distribution samples by investigating the vicinity distributions associated with in-distribution samples. Notably, existing work in this area does not address the more complex scenario of training on non-IID data that includes both in- and out-of-distribution samples.

\subsection{Outlier Detection}
Outlier detection aims to identify and eliminate training samples that diverge significantly from the majority of the data, thereby enhancing the efficacy of downstream learning tasks~\cite{ODD:21}. For example, Iterative Learning (IL)~\cite{IT:18}, which builds upon the LOF algorithm~\cite{LOF:00,OS:13}, iteratively identifies outlier samples from the training data and refines network training accordingly. In another study, feature coupling techniques~\cite{PangCC21} are employed to identify outliers in non-IID categorical data. Outlier Exposure (OE)~\cite{DBLP:conf/iclr/HendrycksMD19} makes use of auxiliary anomalous data to enhance the performance of deep anomaly detectors by training them against a supplementary dataset comprising outliers. Moreover, Outlier Generation~\cite{DBLP:journals/tnn/RiveraKBS22} features a two-level hierarchical latent space model built using autoencoders and variational autoencoders; this method aims to create synthetic but robust anomalies for training binary classifiers. However, it is important to note that these approaches do not specifically address the scenario where both training and test data exhibit non-IID characteristics and include both in- and out-of-distribution samples.

\subsection{Domain Generalization}
Domain generalization aims to train machine learning models capable of generalizing to previously unseen domains by utilizing data from a variety of known domains during the training process. Several notable methods have been proposed in this area. For instance, Domain-Adversarial Neural Network (DANN)~\cite{DBLP:journals/jmlr/GaninUAGLLML16} employs a gradient reversal layer to minimize domain discrepancies while maximizing task performance. Meta-Learning for Domain Generalization (MLDG)~\cite{DBLP:conf/aaai/LiYSH18} framework leverages meta-learning techniques to enhance the network ability to generalize across multiple domains. Domain-Invariant Representation Learning (DIRL)~\cite{DBLP:journals/tmlr/LeviAK22} aims to identify domain-agnostic features by minimizing the mutual information between the learned features and the domain labels. Conditional Domain Adversarial Network (CDAN)~\cite{DBLP:conf/nips/LongC0J18} goes a step further by accounting for the conditional distribution of labels within each domain, thereby improving generalization performance. METABDRY~\cite{DBLP:journals/tnn/0034SC21} employs a combination of pointer networks, adversarial learning, and meta-learning to tackle the challenges posed by sparse boundary tags and a variable output vocabulary. However, it is worth noting that these domain generalization techniques operate under the assumption that all training samples are accurately annotated, which contrasts with the non-IID scenarios we consider, where some training samples may contain incorrect labels.

\subsection{Learning with Noisy Labels}
Learning with noisy labels aims to develop robust models that can be trained effectively even when some of the training samples have incorrect in-distribution labels~\cite{NL:13,AP:19}. This stands in contrast to our focus on non-IID scenarios, where the training data includes out-of-distribution samples that are incorrectly annotated as in-distribution. Various methods exist for tackling the issue of noisy labels. Data cleaning techniques~\cite{DC:16} identify and correct samples with label noise. MentorNet~\cite{MN:18} employs a pre-trained auxiliary network to guide the selection of clean samples during training. Decoupling~\cite{CT:18} and Co-teaching~\cite{DC:17} both utilize dual networks; Decoupling updates each network based on samples that yield differing predictions, whereas Co-teaching trains each network on a subset of samples with low loss, as chosen by the other network. Another avenue of research explores noise-tolerant loss functions. For instance, Mean-Absolute Error (MAE)~\cite{MAE:17} is a symmetric loss that has been proven robust against various types of label noise. Symmetric Cross-Entropy Learning (SL)~\cite{SL:19} enhances traditional cross-entropy loss by adding a noise-robust reverse cross-entropy term, thus making it more symmetric and robust. Bootstrapping~\cite{BT:15} avoids the need for explicit noise modeling by convexly combining the original training labels with the network current predictions. Self-Reweighting from Class Centroids (SRCC)~\cite{DBLP:journals/tnn/MaWYY22} dynamically adjusts the contribution of each sample based on its proximity to class centroids learned online.

\subsection{Other Related Areas}
The non-IID scenarios considered in this study also intersect with several cutting-edge research areas. For instance, open-set recognition~\cite{OSR:20} accounts for out-of-distribution samples by categorizing them into an additional class during training. However, unlike our approach, open-set recognition explicitly identifies the distribution of each training sample. Scene graphs~\cite{DBLP:journals/pami/ChangR00C023} provide a semantic understanding of what is normal or expected in a given type of scene. If an object or relationship appears that does not fit into the established scene graph, it could potentially be flagged as an out-of-distribution item. Unsupervised machine translation~\cite{DBLP:journals/pami/LiHCHYH23} improves translation performance by incorporating multi-modal information from visual content and ensure the trained model can generalize to a different form of data that was not seen during training. Zero-shot learning~\cite{DBLP:journals/pami/ZhangCLLLYH23, DBLP:journals/pami/YanCLGGZZ22} transfer knowledge from seen to unseen activities is similar to identifying out-of-distribution examples based on learned in-distribution data. Specifically, zero-shot temporal activity detection~\cite{DBLP:journals/pami/ZhangCLLLYH23} aims to detect activities that have never been seen during training, and ZeroNAS~\cite{DBLP:journals/pami/YanCLGGZZ22} conducts a differentiable generative adversarial networks architecture search in a specially designed search space for zero-shot learning.

The GL method incorporates elements of Curriculum Learning~\cite{CL:09} and Negative Learning~\cite{NL:19}. Curriculum Learning, as established by previous works~\cite{CL:09}, begins with the network learning simpler aspects of a task and progressively incorporates more challenging examples. In a similar vein, Self-Paced Learning (SPL)~\cite{SPL:10,SPL:14} integrates curriculum design directly into the learning process. Unlike traditional supervised learning methods that update networks based on labeled samples, SPL allows the network itself to dynamically dictate the pace of the curriculum, ranging from simpler to more complex samples. In contrast to these traditional methods, Negative Learning employs complementary labels to reduce the likelihood of erroneous label information affecting the network training. This approach provides a counterpoint to the more standard techniques of supervised learning. On a theoretical level, we employ classical generalization bounds, specifically those based on the Rademacher complexity of a hypothesis class~\cite{RC:02}. The Rademacher complexity serves as a measure of uniform convergence rates. Supporting this theoretical underpinning, Golowich et al.~\cite{SIC:18} have provided bounds on the Rademacher complexity for neural networks, assuming norm constraints on the parameter matrix of each layer. Additionally, we draw inspiration from domain adaptation theories, particularly in the use of divergence measures~\cite{DA:10} to analyze the discrepancy between in-distribution and out-of-distribution samples.

\begin{figure*}
  \centering
  % Requires \usepackage{graphicx}
  \includegraphics[width=1.0\linewidth]{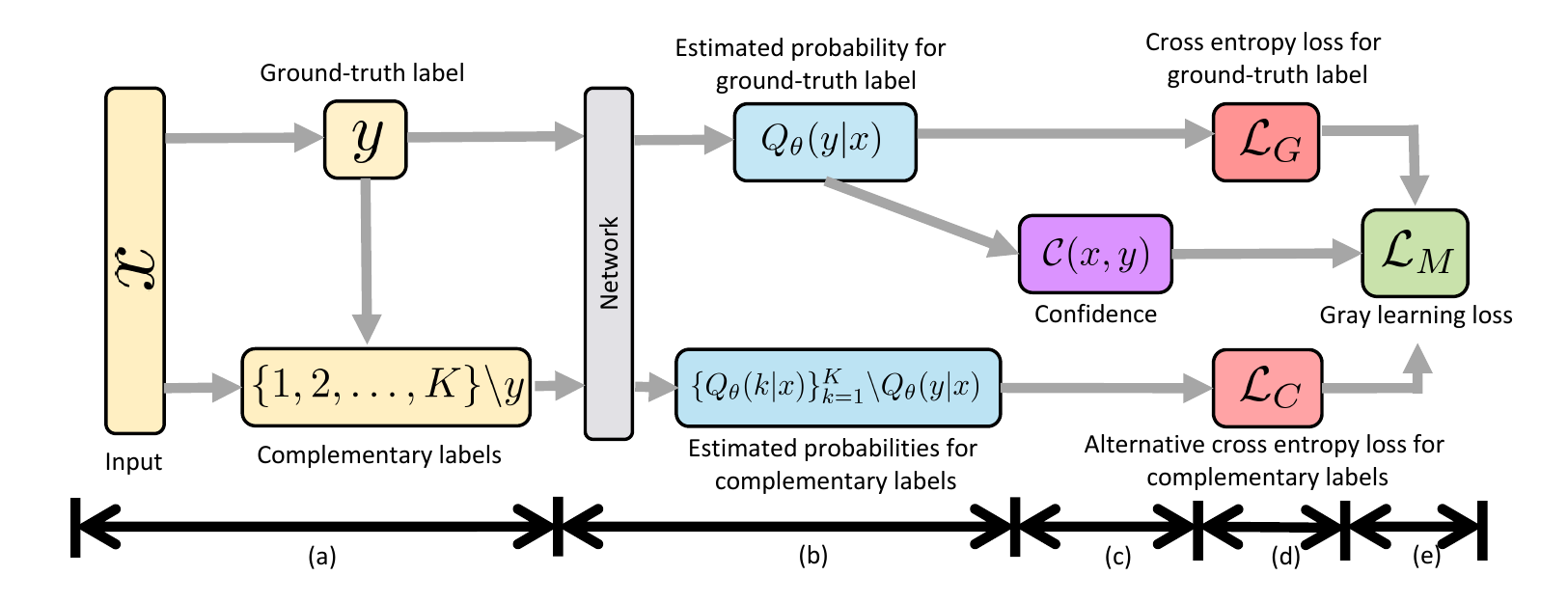}\\
  \caption{The framework of gray learning. (a): Input $x$ corresponds to the ground-truth label $y$ and the complementary labels $\{1,2,\ldots,K\} \backslash y$. The complementary labels of a training sample are inferred from its ground-truth label, which indicate the classes the training sample does not belong to. (b): The network outputs the probabilities $Q_{\theta}(y|x)$ and $\{ Q_{\theta}(k|x) \}_{k = 1}^K \backslash Q_{\theta}(y|x)$ for the corresponding labels. (c):The confidence $\mathcal{C}(x,y)$ is based on the probability of the ground-truth label. (d): The loss functions $\mathcal{L}_{G}$ and $\mathcal{L}_{C}$ are for the ground-truth label and the complementary labels, respectively. We obtain the loss function $\mathcal{L}_{M}$ of gray learning by using $\mathcal{C}(x,y)$ to adaptively adjust the weights for $\mathcal{L}_{G}$ and $\mathcal{L}_{C}$ where a sample with higher confidence provides a higher weight for $\mathcal{L}_{G}$, or vice versa.}
  \label{fig:gl}
\end{figure*}

\section{Gray Learning} \label{sec:tpm}
In the specific scenario we examine, the training dataset is characterized by a non-IID amalgamation of both in-distribution and out-of-distribution samples. In contrast, the test dataset is exclusively composed of in-distribution samples. It is crucial to note that the out-of-distribution samples within the training data are semantically divergent from the in-distribution samples, indicating that they do not align with any recognized in-distribution class. While in-distribution samples carry accurate and reliable annotations, the out-of-distribution samples are misleadingly labeled, as though they belong to in-distribution categories. This distinction underscores the reliability of the labels for in-distribution samples, in stark contrast to the dubious nature of the labels attached to out-of-distribution samples.

As depicted in Fig.~\ref{fig:gl}, GL method employs a phased approach to segregate and address in-distribution and out-of-distribution samples distinctively during the training regimen. Specifically, for each training instance, the GL algorithm performs the following steps:
\begin{enumerate}
    \item Calculates a prediction confidence score, which serves as an estimate of the likelihood that the sample belongs to either the in- or out-of-distribution category.
    \item Assesses loss functions based on both the ground-truth labels and their complementary counterparts.
    \item Dynamically adjusts the weights assigned to the two types of losses in accordance with the prediction confidence score.
\end{enumerate}
In this manner, GL is inclined to rely more heavily on ground-truth labels for samples that exhibit high confidence scores, and on complementary labels for those with low confidence scores. Generally speaking, samples with high confidence are likely to be in-distribution given that such samples form the majority of the training set. Conversely, samples manifesting low confidence scores are considered ambiguous in the initial stages of training but are increasingly likely to be categorized as out-of-distribution as the model evolves.

\subsection{Setup}
Let $\mathcal{X}$ represent the input space and $\mathcal{Y}$ the label space. We define $\mathcal{Y}$ as a finite set containing $K$ classes, formally $\mathcal{Y} = {1, 2, \ldots, K}$. Our training dataset, denoted as $\mathcal{D}$, consists of $N$ samples: $\mathcal{D} = \{ (x_i,y_i)\}_{i = 1}^{N}$. These samples are drawn from an unknown distribution $\mathcal{P}$, which is defined over the Cartesian product $\mathcal{X} \times \mathcal{Y}$.

For a given loss function $\mathcal{L}: \mathcal{Y} \times \mathcal{Y} \rightarrow \mathbb{R}_+$, and any function $h: \mathcal{X} \rightarrow \mathcal{Y}$ that belongs to the hypothesis class $\mathcal{H}$~\cite{CO:08}, we introduce the expected risk:
\begin{equation}
\begin{aligned}
\epsilon_{\mathcal{P}}(\mathcal{L}, h) = \mathbb{E}_{(x,y) \sim \mathcal{P}} \mathcal{L}(h(x),y),
\end{aligned}
\end{equation}
and the empirical risk:
\begin{equation}
\begin{aligned}
\widehat{\epsilon}_{\mathcal{P}}(\mathcal{L}, h) = \frac{1}{N}\sum_{(x,y) \sim \mathcal{D}}  \mathcal{L}(h(x_i),y_i).
\end{aligned}
\end{equation}

We specify the distribution of in-distribution samples as $\mathcal{P}_I$ and that of out-of-distribution samples as $\mathcal{P}_O$. Then, the overall distribution $\mathcal{P}_M$ of the training dataset can be modeled as a weighted mixture of these two distributions:
\begin{equation}
\mathcal{P}_M = (1 - \alpha)\mathcal{P}_I + \alpha \mathcal{P}_O
\end{equation}
where $\alpha \in [0,1]$ serves as a component parameter that controls the proportion of out-of-distribution samples in $\mathcal{P}_M$. We denote an in-distribution sample as $(x^I, y^I) \sim \mathcal{P}_I$ and an out-of-distribution sample as $(x^O, y^O) \sim \mathcal{P}_O$. The training dataset $\mathcal{D}_M$ can be considered as a union of $N_I$ in-distribution and $N_O$ out-of-distribution samples:
\begin{equation}
\mathcal{D}_M = \mathcal{D}_I \bigcup \mathcal{D}_O
\end{equation}
where $\mathcal{D}_I = \{ (x_i^I,y_i^I)\}_{i = 1}^{N_I}$ and $\mathcal{D}_O = \{ (x_i^O,y_i^O)\}_{i = 1}^{N_O}$. It is important to note that the total number of samples $N$ is the sum of $N_I$ and $N_O$, i.e., $N_I + N_O = N$.

Building upon the concept of discrepancy between source and target domains~\cite{DA:10}, we introduce a metric $d_{\mathcal{H}}^{\mathcal{L}}(\mathcal{P}_I, \mathcal{P}_O)$ to quantify the \textit{discrepancy} between in-distribution and out-of-distribution sample distributions. This is formalized as:
\begin{equation}
d_{\mathcal{H}}^{\mathcal{L}}(\mathcal{P}_I, \mathcal{P}_O) = \sup_{h \in \mathcal{H}} \left( \vert \epsilon_{\mathcal{P}_I}(\mathcal{L}, h) - \epsilon_{\mathcal{P}_O}(\mathcal{L}, h)\vert\right).
\label{eq:dc}
\end{equation}
In essence, a low discrepancy value suggests that a hypothesis $h \in \mathcal{H}$ performs comparably under both $\mathcal{P}_I$ and $\mathcal{P}_O$, and vice versa. For the mixture distribution $\mathcal{P}_M$, we estimate the conditional distribution $\mathcal{P}_M(y | x)$ by a parameterized distribution $Q_{\theta}(y | x)$ with model parameter $\theta$. The softmax value of label $y$ is formulated as:
\begin{equation}
Q_{\theta}(y | x) = \frac{\exp{h_{\theta}^y(x)} }{ \sum_{k \in [K]} \exp{h_{\theta}^k(x)}}.
\end{equation}
In the context of deep learning, the hypothesis $h_{\theta}$ corresponds to a parameterized neural network. $h_{\theta}(x) = \{ h_{\theta}^k(x)\}_{k = 1}^K$ gives the network output for an input $x$.

\subsection{Objective Function}
The algorithmic procedure for the GL method is further delineated in Algorithm~\ref{alg:GL}.
\subsubsection{Confidence Calculation}
Unlike Maximum Softmax Probability (MSP), a conventional method for out-of-distribution detection that calculates confidence scores for unlabeled samples during the testing phase, GL differs by focusing on labeled samples within the training phase. In essence, while softmax distributions of MSP serve as a basis for discerning in-distribution versus out-of-distribution samples, GL adapts this approach to suit its unique training context.

Specifically, we modify the confidence calculation methodology of MSP to take into account ground-truth labels. The confidence of a training input $x$ with its associated label $y$ is then defined as follows:
\begin{equation}
\mathcal{C}(x,y) = Q_{\theta}(y | x).
\label{eq:cf}
\end{equation}

Given that in-distribution samples constitute the majority of the training dataset, a sample $(x,y)$ with a high confidence score is more likely to be an in-distribution sample. However, it is important to note that a low-confidence score does not definitively categorize a sample as either in-distribution or out-of-distribution. This is because, during the early stages of training, all samples, regardless of their true distribution, tend to exhibit low confidence scores.

\subsubsection{Cross Entropy Loss for Ground-truth Labels}
For a sample $(x,y)$ that exhibits high confidence and is thus likely to be an in-distribution instance, we employ the standard cross-entropy loss $\mathcal{L}_{G}$. This loss function maps the input $x$ to its corresponding ground-truth label $y$, mathematically represented as,
\begin{equation}
\mathcal{L}_{G}(x,y) =  - \log Q_{\theta}(y | x).
\label{eq:LH}
\end{equation}
For samples that are truly in-distribution, this cross-entropy loss $\mathcal{L}_{G}$ is effective because the ground-truth labels are reliable indicators of the actual classes. However, when it comes to out-of-distribution samples, the application of $\mathcal{L}_{G}$ can be detrimental. This is because these samples are erroneously tagged with labels from in-distribution classes, to which they do not actually belong. Consequently, $\mathcal{L}_{G}$ ill-suited for a training dataset comprising a non-IID mixture of in- and out-of-distribution samples, especially if the objective is to train a robust network.

\subsubsection{Alternative Cross Entropy Loss for Complementary Labels}
For low-confidence samples $(x,y)$, the determination of whether they belong to in- or out-of-distribution classes is ambiguous, especially in the early stages of training when most samples exhibit low confidence. To address this, we introduce the concept of complementary labels. These labels are useful for both in- and out-of-distribution samples, as neither set belongs to the classes represented by these complementary labels.

For a given training input $x$ with its associated label $y$, we define the corresponding set of complementary labels as:
\begin{equation}
\mathcal{Z}(x,y) = \{1,2,\ldots,K\} \backslash y.
\end{equation}
Subsequently, we employ an alternative cross-entropy loss $\mathcal{L}_{C}$, designed to negate the corresponding complementary labels. This loss function is mathematically formulated as
\begin{equation}
\mathcal{L}_{C}(x,\mathcal{Z}(x,y)) =  - \sum_{y' \in \mathcal{Z}(x,y) }\log \left( 1 - Q_{\theta}(y' | x) \right).
\label{eq:LL}
\end{equation}
For an in-distribution sample, $\mathcal{L}_{C}$ helps identify the ground-truth label by actively excluding the complementary labels. While out-of-distribution samples can also benefit from this approach by improving their confidence on the misleading ground-truth label and excluding the complementary labels. Accordingly, $\mathcal{L}_{C}$ is less risky than $\mathcal{L}_{G}$ in terms of providing incorrect label information, as it does not directly map inputs to ground-truth labels.

\subsubsection{Adaptively Weighting Loss}
GL obviates the need for selecting in-distribution samples based on confidence during training. Unlike traditional methods, GL capitalizes on the use of complementary labels, allowing it to learn effectively from a non-IID dataset comprising both in- and out-of-distribution samples.

For high-confidence samples, GL applies a higher weight to the cross-entropy loss $\mathcal{L}_{G}$, which targets ground-truth labels., thereby widening the confidence gap between in- and out-of-distribution samples. Conversely, for low-confidence samples, GL emphasizes the alternative cross-entropy loss $\mathcal{L}_{C}$, which targets complementary labels. This allows the model to glean accurate label information even when the ground-truth labels may be untrustworthy.

To balance these objectives, GL employs a weighted loss function $\mathcal{L}_{M}(x,y)$, defined as follows:
\begin{equation}
\begin{aligned}
\mathcal{L}_{M}(x,y) = &  \mathcal{C}(x,y) \mathcal{L}_{G}(x,y)  \\
& +  (1 - \mathcal{C}(x,y))\mathcal{L}_{C}(x,\mathcal{Z}(x,y)).
\label{eq:LA}
\end{aligned}
\end{equation}
Here, $\mathcal{C}(x,y)$ serves as the weighting factor, derived from the prediction confidence according to Eq.~(\ref{eq:cf}). The expected risk under this loss function $\epsilon_{\mathcal{P}_M}(\mathcal{L}_M, h)$ is then given by
\begin{equation}
\epsilon_{\mathcal{P}_M}(\mathcal{L}_M, h) = \int \mathcal{L}_{M}(x,y) \mathrm{d} \mathcal{P}_M.
\label{eq:OB}
\end{equation}
To empirically estimate this risk, we obtain $N$ samples from the mixture distribution $\mathcal{P}_M$ which includes $N_I$ in-distribution samples and $N_O$ out-of-distribution samples. The empirical risk of the expected risk $\mathcal{L}_{M}(x,y)$ is defined as:
\begin{equation}
\widehat{\epsilon}_{\mathcal{P}_M}(\mathcal{L}_M, h) =  \frac{1}{N} \sum_{(x,y) \sim \mathcal{D}_M} \mathcal{L}_{M}(x,y).
\label{eq:OB2}
\end{equation}

As the training progresses, GL dynamically adjusts the weighting between $\mathcal{L}_{G}$ and $\mathcal{L}_{C}$ based on the evolving confidence levels of the samples. Specifically, when the confidence level of a sample $(x,y)$ is low, GL focuses on learning from the complementary labels, as it is uncertain whether the sample is in- or out-of-distribution. Once the confidence level rises, indicating likely in-distribution status, GL shifts its focus towards the ground-truth label for more direct and effective learning. This iterative strategy enables GL to increasingly segregate in- and out-of-distribution samples as training progresses, culminating in a network that is robust and versatile.

\begin{algorithm*}[t]
    \caption{Gray Learning}
    \label{alg:GL}
    \begin{algorithmic}[1]
    \STATE {\bfseries Input:} training dataset $\mathcal{D}_M$\\
    \REPEAT
    \FOR{ $(x,y)$ in  $\mathcal{D}_M$}
        \STATE Obtain its complementary label set: $\mathcal{Z}(x,y) = \{1,2,\ldots,K\} \backslash y$
        \STATE Calculate the confidence: $\mathcal{C}(x,y) = Q_{\theta}(y | x)$
        \STATE Estimate the loss for the ground-truth label: $\mathcal{L}_{G}(x,y) = - \log Q_{\theta}(y | x)$
        \STATE Estimate the loss for complementary labels: $\mathcal{L}_{C}(x,\mathcal{Z}(x,y)) =  - \sum_{y' \in \mathcal{Z}(x,y) }\log \left( 1 - Q_{\theta}(y' | x) \right)$
        \STATE Obtain the adaptively weighting loss
        \begin{equation*}
            \mathcal{L}_{M}(x,y) = \mathcal{C}(x,y) \mathcal{L}_{G}(x,y)  +  (1 - \mathcal{C}(x,y))\mathcal{L}_{C}(x,\mathcal{Z}(x,y))
        \end{equation*}
    \ENDFOR
    \STATE Estimate the empirical risk $\widehat{\epsilon}_{\mathcal{P}_M}(\mathcal{L}_M, h_{\theta}) $
    \STATE Obtain gradients $\nabla_{\theta} \widehat{\epsilon}_{\mathcal{P}_M}(\mathcal{L}_M, h_{\theta})$ to update parameters $\theta$
    \UNTIL{convergence}
    \STATE {\bfseries Output:} parameterized network $h_{\theta}$
\end{algorithmic}
\end{algorithm*}

\section{Theoretical Guarantees} \label{sec:the}
In this section, we present the theoretical results that guarantee the efficacy of the GL method. Unlike standard methods that solely optimize the conventional cross-entropy loss, GL is designed to work with non-IID datasets that include a mixture of both in-distribution and out-of-distribution samples. Specifically, in the training phase, in-distribution samples come with reliable annotations, while out-of-distribution samples are erroneously treated as in-distribution and are labeled as such.

We juxtapose GL against a baseline approach herein referred to as the standard method. The standard method employs the traditional cross-entropy loss $\mathcal{L}_G$ for optimization, while GL uses the weighted loss $\mathcal{L}_M$. We define the hypothesis that minimizes the loss function for the standard method on the mixed distribution $\mathcal{P}_M$ of in- and out-of-distribution samples as $\widehat{h}_{M}$. Analogously, the hypothesis that minimizes the loss function for GL under the same distribution is defined as $\widetilde{h}_M$. They are:
\begin{equation}
\begin{aligned}
\widehat{h}_{M} & = \arg \min_{h \in \mathcal{H}} \widehat{\epsilon}_{\mathcal{P}_M}(\mathcal{L}_G, h), \\
\widetilde{h}_M & = \arg \min_{h \in \mathcal{H}} \widehat{\epsilon}_{\mathcal{P}_M}(\mathcal{L}_M, h).
\end{aligned}
\end{equation}
It is worth noting that the test data comprise exclusively in-distribution samples. Therefore, both $\widehat{h}_{M}$ from the standard method and $\widetilde{h}_M$ from GL aspire to approximate the optimal hypothesis $h_{I}^*$ tailored for the in-distribution sample. The optimal hypothesis $h^*_I$ is formally defined as:
\begin{equation}
h^*_I = \arg \min_{h \in \mathcal{H}} \epsilon_{\mathcal{P}_I}(\mathcal{L}_G, h).
\end{equation}
With this theoretical framework, we set the stage for further discussions and proofs, which will quantify the relative advantages of using the GL method in specific scenarios involving non-IID data.

\subsection{Generalization Error of Standard Method.}
In the context of a neural network trained via the standard method over a mixed distribution $\mathcal{P}_M$ comprising both in-distribution and out-of-distribution samples, the following theorem establishes an upper bound for the expected risk when evaluated on the target in-distribution $\mathcal{P}_I$. For a more detailed mathematical derivation, readers are referred to Appendix~\ref{tm:rb}.
\begin{theorem}
Assume (1) $\mathcal{H}$ is the class of real-valued networks of depth $d$ over the domain $\mathcal{X}$, and $x \in \mathcal{X}$ is upper bounded by $B$, i.e., for any $x$, $\Vert x \Vert \leq B$; (2) the Frobenius norm of the weight matrices $W_1,\ldots,W_d$ are at most $M_1,\ldots,M_d$; (3) the loss function $\mathcal{L}$ is $L$-Lipschitz continuous w.r.t. $h \in \mathcal{H}$ and $\vert \mathcal{L}(h, y)\vert \leq c$ for all $y$ and $h \in \mathcal{H}$; (4) the activation function is 1-Lipschitz, positive-homogeneous, and applied element-wise (such as the ReLU). For any $\delta > 0$, with probability at least $1 - \delta$, we have
\begin{equation*}
\begin{aligned}
& \epsilon_{\mathcal{P}_I}(\mathcal{L}_G, \widehat{h}_M) - \epsilon_{\mathcal{P}_I}(\mathcal{L}_G, h_I^*)\\
\leq &  2 \alpha d_{\mathcal{H}}(\mathcal{P}_I, \mathcal{P}_O)\\
+ &  \frac{4BL(\alpha \sqrt{N_I} + (1 - \alpha) \sqrt{N_O})}{\sqrt{N_I N_O}} (\sqrt{2d \ln2} + 1) \prod_{i = 1}^d M_i\\
+ & \frac{8c(\alpha\sqrt{N_I} + (1-\alpha)\sqrt{N_O})}{\sqrt{N_I N_O}}\sqrt{2 \ln(16 / \delta)}.
\end{aligned}
\end{equation*}
\label{tm:rb}
\end{theorem}
In analyzing the performance trade-offs, we identify three contributing terms that influence the gap between the minimizer $\widehat{h}_{M}$ of the empirical risk $\widehat{\epsilon}_{\mathcal{P}_M}(\mathcal{L}_G, h)$ on the non-IID samples and the optimal $h^*_I$ of the expected risk $\epsilon_{\mathcal{P}_I}(\mathcal{L}_G, h)$ on the in-distribution samples. The first term addresses the intrinsic distributional differences between in-distribution and out-of-distribution samples. The coefficient $\alpha$ quantitatively signifies this impact. Intuitively, a higher prevalence of out-of-distribution samples in the training data will degrade the in-distribution classification performance, aligning with our empirical expectations. The second term is crucial and pertains to the characteristics of both the neural network architecture and the cross-entropy loss function being used. It encapsulates how well the model can generalize from the training data to unseen in-distribution samples. The impact of this term can vary based on hyperparameter settings, network depth, and other architectural nuances. The third term is associated with the stochastic nature of the training data sampled from $\mathcal{P}_M$. A larger dataset can mitigate the sampling bias, thereby reducing the gap between $\widehat{h}_{M}$ and $h^*_I$.

\subsection{Equivalent form of Gray Learning}
In order to derive the generalization error for our proposed GL method, we reframe the objective function, as expressed in Eq.~(\ref{eq:OB}), into an equivalent formulation. The comprehensive derivation of this equivalent form is available in Appendix~\ref{tm:ob}.
\begin{theorem}
For training samples drawn from the mixture distribution $\mathcal{P}_M$, the expected risk of GL can be rewritten as:
\begin{equation*}
\begin{aligned}
& -\int \log \mathcal{L}_G(x,y) \mathrm{d} P_M,\\
\text{s.t.~} & r(\theta, x, y) \leq \lambda, \forall (x,y) \sim \mathcal{P}_M,\\
\end{aligned}
\end{equation*}
where
\begin{equation*}
\begin{aligned}
r(\theta, x, y) = & (1 - Q_{\theta}(y|x)) \log Q_{\theta}(y|x) \left( 1 - Q_{\theta}(y|x) \right) \\
 - & (1 - Q_{\theta}(y|x)) \sum_{k = 1}^K \log \left( 1 - Q_{\theta}(y|x) \right),
\end{aligned}
\end{equation*}
and $\lambda > 0$.
\label{tm:ob}
\end{theorem}
We observe that the objective function employed by the GL method can be construed as a constrained form of the conventional cross-entropy loss, where each training sample $(x, y)$ is subjected to a regularizer $r(\theta, x, y)$ capped by an upper bound $\lambda$. This parameter $\lambda$ dynamically modulates the influence of the regularizer during the training process, allowing for more nuanced learning behavior. Interestingly, as $\lambda$ tends toward infinity, the GL method reverts to the baseline approach, effectively neutralizing its specialized handling of in-distribution and out-of-distribution samples.

\subsection{Generalization Error of Gray Learning}
For a network trained using the GL method on the mixed distribution $\mathcal{P}_M$ encompassing both in-distribution and out-of-distribution samples, the ensuing theorem delineates a theoretical upper bound on the expected risk for the target in-distribution $\mathcal{P}_I$. Additionally, the theorem specifies conditions under which the GL method outperforms the standard approach. The full derivation of this theorem can be found in Appendix~\ref{tm:gegl}.
\begin{theorem}
Following the conditions of Theorem~\ref{tm:rb}, let $\log \sum_{k = 1}^K \exp(h_{\theta}^k(x)) \leq z$ for any $x$. For any $\delta > 0$, with probability at least $1 - \delta$, we have
\begin{equation*}
\begin{aligned}
& \epsilon_{\mathcal{P}_I}(\mathcal{L}_G, \widetilde{h}_M) - \epsilon_{\mathcal{P}_I}(\mathcal{L}_G, h_I^*) \\
\leq &  2 \alpha d_{\mathcal{H}}(\mathcal{P}_I, \mathcal{P}_O)\\
+ &  \frac{4BLK(\alpha \sqrt{N_I} + (1 - \alpha) \sqrt{N_O})}{\sqrt{N_I N_O}} (c + \log(2\lambda - 2))\\
+ & \frac{8c(\alpha\sqrt{N_I} + (1-\alpha)\sqrt{N_O})}{\sqrt{N_I N_O}}\sqrt{2 \ln(16 / \delta)}.
\end{aligned}
\end{equation*}
This bound is tighter than that of the baseline method if
\begin{align*}
\lambda \leq 1 + \frac{1}{2} \exp{\left( \frac{B(\sqrt{2d \ln2} + 1) \prod_{i = 1}^d M_i}{L\sqrt{K}} - z \right)}.
\end{align*}
\label{tm:gegl}
\end{theorem}
It is noteworthy that the performance disparity between the optimal hypothesis $\widetilde{h}_M$, which minimizes the empirical risk under GL for non-IID data, and $h^*_I$, the minimizer of the expected risk for in-distribution samples, is conceptually akin to the gap elucidated for the standard method in Theorem~\ref{tm:rb}. The first term of this gap is attributed to the unavoidable distribution discrepancy between in-distribution and out-of-distribution samples, as captured by Eq.~(\ref{eq:dc}). The divergence in the second term arises from the contrasting empirical risks between the standard and GL methods. Intriguingly, if the regularizer $r(\theta, x, y)$ adheres to the criteria outlined in Theorem~\ref{tm:gegl}, GL exhibits robust learning capabilities even when faced with non-IID data. The parameter $\lambda$ serves as an implicit, adaptively-adjusted variable throughout the training regimen. Owing to their ability to effectively model complex data structures, deep neural networks can iteratively minimize $\lambda$ to fulfill the stated condition in real-world scenarios.

\begin{table*}[t]
  \renewcommand{\arraystretch}{1.3}
  \setlength\tabcolsep{4pt}
  \centering
  \caption{Classification accuracy on imagery data. All values are in percentage, and boldface values show the relatively better classification performance.}
  \label{tb:cp}
\begin{tabular}{cccccccccccc}
\hline
In-distribution          & Out-of-distribution label & Standard & MAE & BT  & SPL & IL  & SL & NL & SRCC & LCVD & GL  \\ \hline \hline
\multirow{2}{*}{CIFAR10} & Specific                  & 93.8 & 94.1 & 95.1 & 89.9 & 94.8 & 94.7 &94.8 & 94.7 & 94.2 & \textbf{95.2}\\ %\cline{2-10}
                         & Random                    & 93.7 & 94.3 & 95.1 & 90.9 & 95.1 & 94.9 &94.5 & 94.7 & 94.4 & \textbf{95.3}\\ \hline \hline
\multirow{2}{*}{SVHN}    & Specific                  & 95.0 & 96.3 & 96.0 & 92.0 & 96.1 & 96.4 &96.1 & 96.4 & 95.2 & \textbf{96.7}\\ %\cline{2-10}
                         & Random                    & 95.2 & 96.7 & 96.4 & 93.5 & 96.5 & 96.6 &96.4 & 96.2 & 95.3 & \textbf{96.9}\\ \hline \hline
\multirow{2}{*}{CIFAR100}& Specific                  & 66.3 & 69.9 & 67.6 & 69.2 & 66.8 & 70.8 & 70.3 & 70.3 & 71.1 &\textbf{77.6}\\ %\cline{2-10}
                         & Random                    & 67.1 & 69.3 & 67.5 & 69.2 & 67.1 & 70.7 & 70.5 & 71.2 & 72.8 &\textbf{77.2}\\ \hline
%\multicolumn{2}{c}{Average}                          & 0 & \multicolumn{1}{l}{95.33} & \multicolumn{1}{l}{95.66} & \multicolumn{1}{l}{91.62} & \multicolumn{1}{l}{95.66} & \multicolumn{1}{l}{95.67} & \multicolumn{1}{l}{95.51} & \multicolumn{1}{l}{\textbf{96.04}} \\ \hline
\end{tabular}
\end{table*}

\begin{table*}[t!]
  \renewcommand{\arraystretch}{1.3}
  \setlength\tabcolsep{4pt}
  \centering
  \caption{Classification accuracy on tabular data. All values are in percentage, and boldface values show the relatively better classification performance.}
  \label{tb:cp2}
\begin{tabular}{ccccccccccc}
\hline
In-distribution & Standard & MAE   & BT    & SPL   & IL    & SL    & NL   & SRCC & LCVD  & GL    \\ \hline \hline
Abalone	        & 79.1 & 81.6 & 80.8 & 49.4 & 80.3 & 80.9 & 79.6 & 80.2 & 80.6 & \textbf{82.1} \\
Arrhythmia	    & 77.9 & 78.5 & 78.3 & 81.0 & 78.6 & 79.8 & 79.2 & 81.5 & 81.2& \textbf{83.4} \\
Gene	        & 39.2 & 38.1 & 38.8 & 38.3 & 37.9 & 35.7 & 40.6 & 41.2 & 42.3& \textbf{44.8} \\
Iris	        & 59.0 & 60.0 & 64.0 & 50.0 & 66.0 & 64.0 & 60.0 & 75.9 & 77.2& \textbf{95.0} \\
Skyserver	    & 68.6 & 68.4 & 68.8 & 68.7 & 68.7 & 68.6 & 68.8 & 72.1 & 74.1& \textbf{78.9} \\
Speech	        & 53.0 & 52.8 & 53.8 & 44.6 & 53.8 & 52.3 & 53.8 & 53.2 & 52.3& \textbf{56.0} \\
Stellar	        & 56.8 & 56.8 & 56.6 & 56.8 & 56.7 & 56.9 & 56.5 & 66.3 & 65.4& \textbf{76.1} \\
WineQT	        & 59.0 & 57.0 & 56.0 & 48.0 & 56.1 & 55.8 & 55.5 & 60.2 & 60.7& \textbf{62.5} \\ \hline \hline
Average         & 61.6 & 61.7 & 62.1 & 54.6 & 62.3 & 61.8 & 61.8 & 64.2 & 66.3& \textbf{72.4} \\ \hline
\end{tabular}
\end{table*}

\section{Experiment Results} \label{sec:exp}
In the absence of directly comparable methods specifically designed for handling non-IID data comprised of both in- and out-of-distribution samples during training, we validate the efficacy of our proposed GL method\footnote{The source code is publicly available at: \url{https://github.com/Lawliet-zzl/GL}.}. For our evaluations, we compare GL against a baseline approach, which we refer to as the standard method, as well as alternative techniques drawn from the field of robust statistics. The standard method exclusively employs traditional cross-entropy loss for optimization and lacks a dedicated mechanism for addressing out-of-distribution samples.

We conduct an exhaustive set of experiments to assess the robustness and versatility of GL. These experiments include an analysis of the effects of varying proportions, sources, and labels of out-of-distribution samples when mixed with in-distribution training samples. Additionally, we extend our evaluation to multiple network architectures, assess the calibration capabilities, and carry out an ablation study to identify key factors contributing to the performance.

\begin{table}[]
\renewcommand{\arraystretch}{1.3}
\center
\caption{Statistics of datasets.}
\label{tb:ds}
\begin{tabular}{ccccc}
\hline
Type                     & Dataset    & \# instances & \# features & \# labels \\ \hline\hline
\multirow{3}{*}{Imagery} & CIFAR10    & 60000        & 1024        & 10        \\ %\cline{2-5}
                         & SVHN       & 99289        & 1024        & 10        \\ %\cline{2-5}
                         & CIFAR100   & 60000        & 1024        & 100       \\ \hline \hline
\multirow{8}{*}{Tabular} & Abalone    & 4177         & 8           & 3         \\ %\cline{2-5}
                         & Arrhythmia & 87553        & 187         & 5         \\ %\cline{2-5}
                         & Gene       & 801          & 20531       & 5         \\ %\cline{2-5}
                         & Iris       & 150          & 150         & 3         \\ %\cline{2-5}
                         & Skyserver  & 100000       & 17          & 3         \\ %\cline{2-5}
                         & Speech     & 3960         & 12          & 6         \\ %\cline{2-5}
                         & Stellar    & 100000       & 16          & 3         \\ %\cline{2-5}
                         & WineQT     & 1143         & 11          & 6         \\ \hline
\end{tabular}
\end{table}

\subsection{Settings}
In our experimental evaluation, we assess performance of GL across both imagery and tabular data, the details of which are summarized in \tablename~\ref{tb:ds}. For the non-IID data, the proportion of out-of-distribution samples is controlled by the component parameter $\alpha$. Unless otherwise stated, we set $\alpha=0.1$ as a default value for our experiments. To generate complementary labels for each training sample, we take all labels other than the ground-truth label as its complementary set. Our evaluation metrics include \textit{Classification Accuracy} for gauging the discriminative capabilities and \textit{Expected Calibration Error} (ECE)~\cite{CAL:17} for measuring the predictive confidence calibration. ECE quantifies the difference between the network predicted confidence and its actual classification accuracy. A well-calibrated model will exhibit high confidence for correctly classified samples and low confidence for misclassifications. In evaluating ECE, we divide the confidence range into $20$ bins to capture detailed calibration behavior across different levels of predictive confidence.

\subsubsection{Imagery Data}
In our experiments, we utilize a diverse range of datasets to create non-IID training data, incorporating both in-distribution and out-of-distribution samples. Specifically, for in-distribution data, we employ CIFAR10~\cite{CIFAR10:09}, SVHN~\cite{SVHN:11}, and CIFAR100~\cite{CIFAR10:09}. We choose Mini-Imagenet~\cite{IMAGENET:09} as our source for out-of-distribution data, using 100 classes unless otherwise specified.

For CIFAR10 and SVHN, each featuring 10 classes, we segment Mini-Imagenet into 10 balanced subsets according to class labels. For CIFAR100, which contains 100 classes, we use all samples from Mini-Imagenet as the out-of-distribution data. We introduce two types of labeling for these out-of-distribution samples: \textit{specific labels} and \textit{random labels}. In the specific-labeling scheme, all out-of-distribution samples from the same Mini-Imagenet class receive the same in-distribution label. In contrast, the random-labeling scheme assigns a randomly selected in-distribution label to each out-of-distribution sample.

We preprocess the training data using standard data augmentation techniques, including resizing, random cropping, and random horizontal flipping. Our evaluation spans six different neural network architectures: ResNet18~\cite{RES:16}, VGG19~\cite{VGG:15}, ShuffleNetV2~\cite{SHU:18}, MobileNetV2~\cite{MON:18}, SqueezeNet~\cite{SE:16}, and DenseNet121~\cite{DEN:17}. Parameters are updated using Stochastic Gradient Descent~\cite{ML:14}, with an initial learning rate of $0.1$, reduced by a factor of $10$ at epochs $100$ and $150$. All models are trained for $200$ epochs with mini-batches of $128$ samples.

\subsubsection{Tabular Data}
We extend our evaluation to include small tabular datasets, encompassing two from the UCI repository\footnote{https://archive.ics.uci.edu/} (Abalone and Iris) as well as six from Kaggle\footnote{https://www.kaggle.com/} (Arrhythmia, Gene, Skyserver, Speech, Stellar, and WineQT). In each case, we designate samples from the smallest class as out-of-distribution and the remaining samples as in-distribution. For these tabular datasets, we employ shallow, fully-connected neural networks with a two-layer architecture featuring $128$ ReLU units in each hidden layer. We use the Adam optimizer~\cite{ADAM:15} with default hyperparameters, training the models over $10$ epochs in mini-batches of $16$ samples.

\begin{figure}
  \centering
  \includegraphics[width=0.48\linewidth]{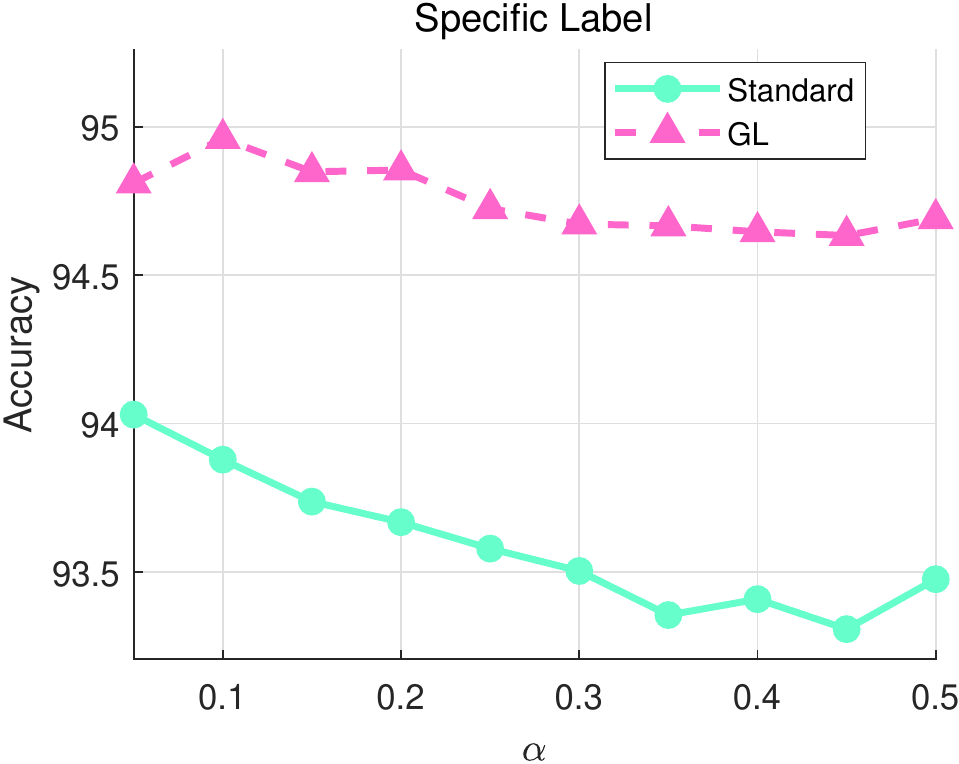}
  \hspace{0.1cm}
  \includegraphics[width=0.48\linewidth]{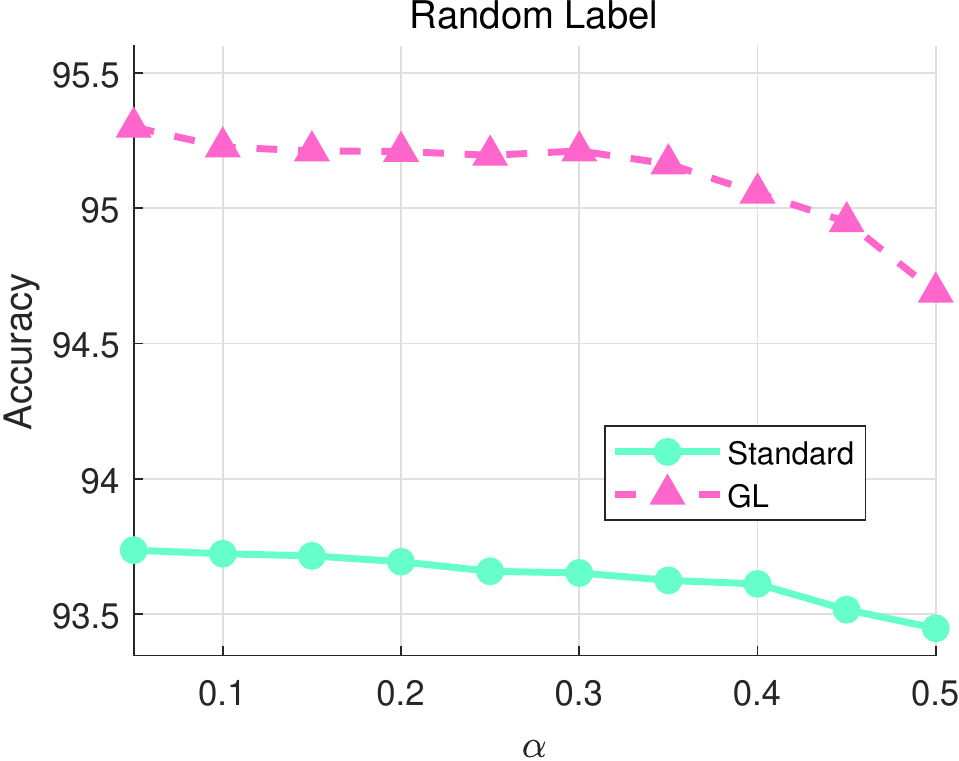}
  \caption{The effect of the proportion of out-of-distribution samples.}\label{fig:ood}
\end{figure}

\begin{figure}
  \centering
  \includegraphics[width=0.48\linewidth]{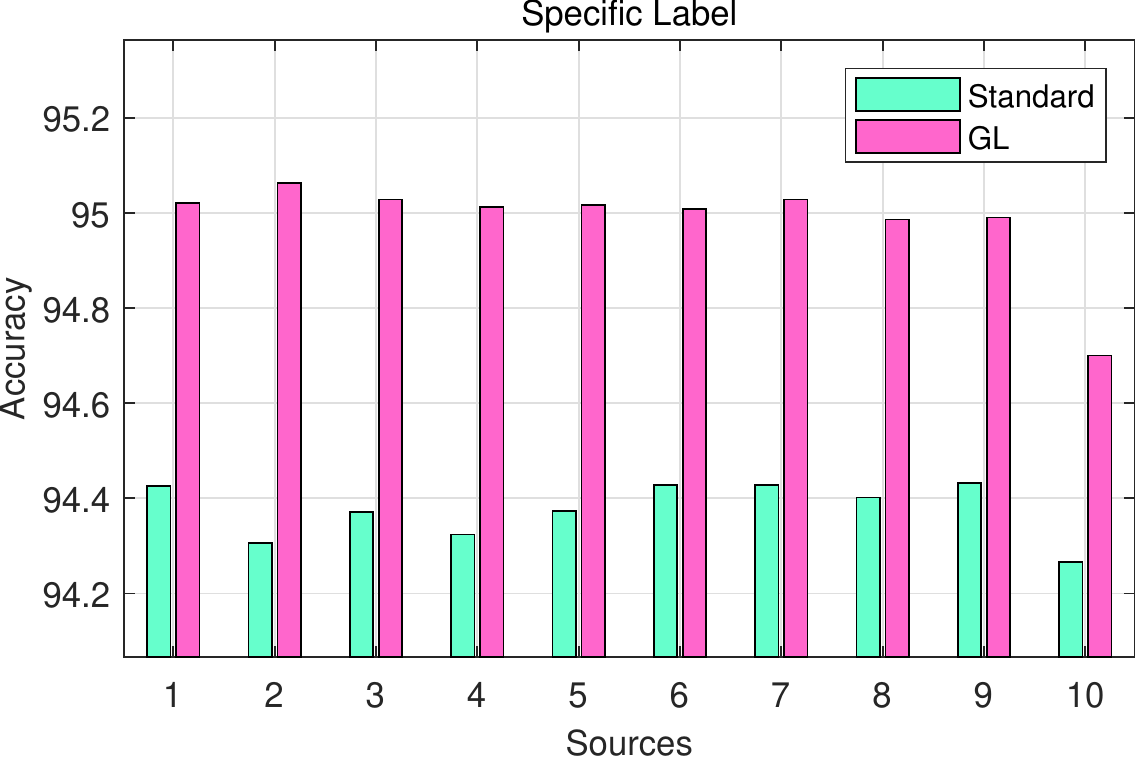}
  \hspace{0.1cm}
  \includegraphics[width=0.48\linewidth]{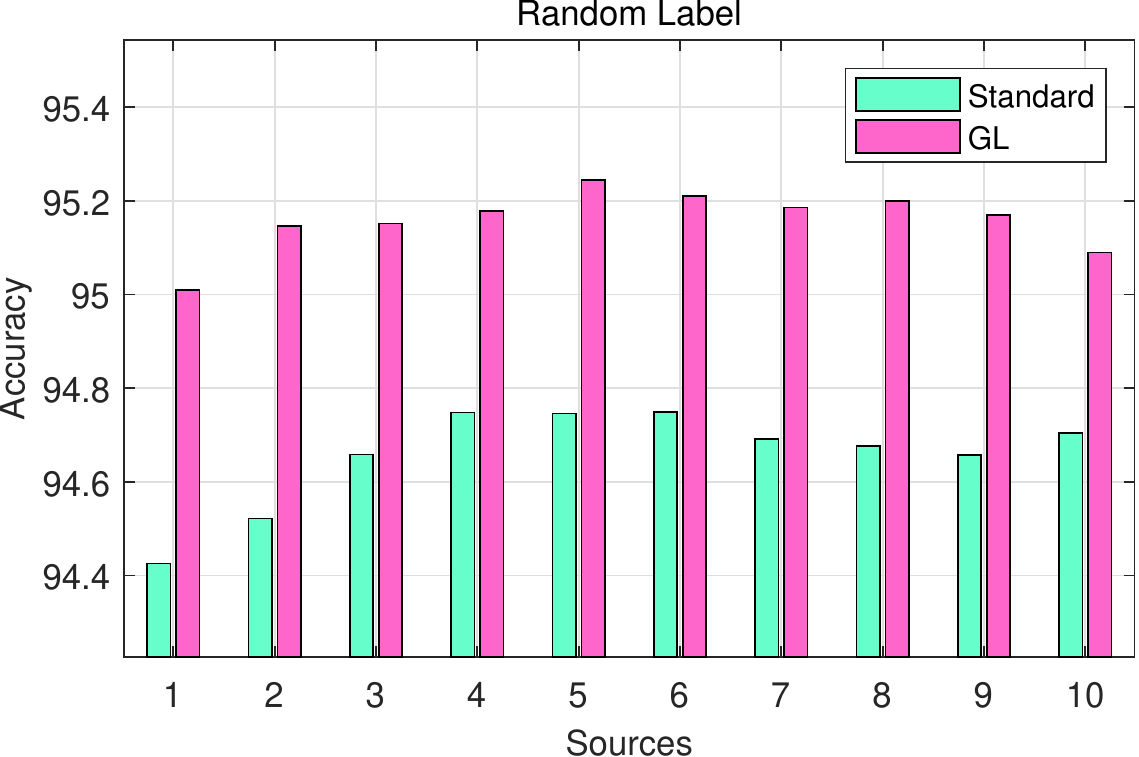}
  \caption{The effect of the out-of-distribution inputs. An index in $\{1, \ldots, 10\}$ represents a source of out-of-distribution inputs.}\label{fig:data}
\end{figure}

\subsection{Comparison Results}
This study pioneers the investigation into non-IID data that encompasses both in-distribution and out-of-distribution samples. Given the absence of direct competitors, we benchmark GL against various alternative methods derived from related research domains. These methods are also suitable for handling the non-IID data under study. Specifically, the alternative methods include Mean-Absolute Error (MAE)~\cite{MAE:17}, Bootstrapping~\cite{BT:15}, Iterative Learning (IL)~\cite{IT:18}, and Symmetric cross entropy Learning (SL)~\cite{SL:19}, Self-Reweighting from Class Centroids (SRCC)~\cite{DBLP:journals/tnn/MaWYY22}, Learning from Cross-class Vicinity Distribution (LCVD)~\cite{LCVD:23}. The other related methods include Self-Paced Learning (SPL)~\cite{SPL:14} and Negative Learning (NL)~\cite{NL:19}. For the image datasets, we deploy the ResNet18 architecture and designate samples from Mini-Imagenet as out-of-distribution. In the case of CIFAR10 and SVHN, we confine ourselves to the first source of Mini-Imagenet to ensure an equivalent number of classes. For CIFAR100, all sources from Mini-Imagenet are incorporated.

The comparison results for imagery data are outlined in \tablename~\ref{tb:cp}. Our findings reveal that Self-Paced Learning (SPL) consistently underperforms in terms of classification accuracy across all examined combinations of in-distribution datasets and out-of-distribution labels. This outcome suggests that SPL is not effective at discerning between in-distribution and out-of-distribution samples during the training process. In contrast, BT and IL show a marked improvement, specifically $4.41 \%$ gains on CIFAR10 and SVHN. These results indicate that strategies that filter out out-of-distribution samples and avoid directly fitting to non-IID data can enhance the robustness of learned models. NL only marginally outperforms the straightforward MAE method by $0.46 \%$, suggesting that merely utilizing complementary labels is not sufficient for capturing the nuances required for accurate classification. Remarkably, our proposed GL method outshines all competitors, achieving a $7.2 \%$ improvement over the worst-performing SPL and a $3.3 \%$ improvement over the next best method, SL. This superiority can be attributed to capability of GL to ameliorate the limitations of both SPL and NL by adaptively reweighting training samples based on their confidence levels and reconsidering the ground-truth labels for high-confidence samples.

The comparative analysis for small tabular data is presented in \tablename~\ref{tb:cp2}. Notably, GL significantly outperforms the competing methods, showing an average classification accuracy improvement of $16.7\%$ across all datasets. Previous studies have highlighted that neural networks are prone to overfitting when dealing with small, unreliable tabular data~\cite{OF:21,ZhuCY22}, leading to poor generalization on test in-distribution samples. GL addresses this issue by adaptively utilizing ground-truth labels for in-distribution samples and complementary labels for out-of-distribution samples. The use of complementary labels mitigates the risk of introducing erroneous label information, enhancing the robustness. Consequently, GL demonstrates superior performance even when applied to small tabular datasets.

\begin{figure}
  \centering
  \includegraphics[width=0.48\linewidth]{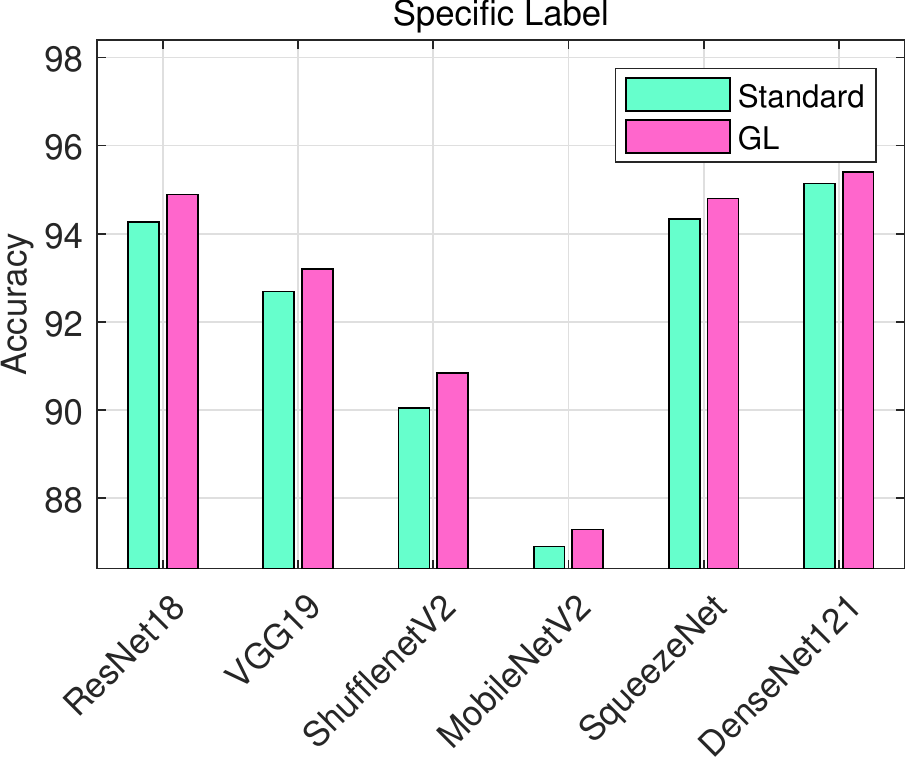}
  \hspace{0.1cm}
  \includegraphics[width=0.48\linewidth]{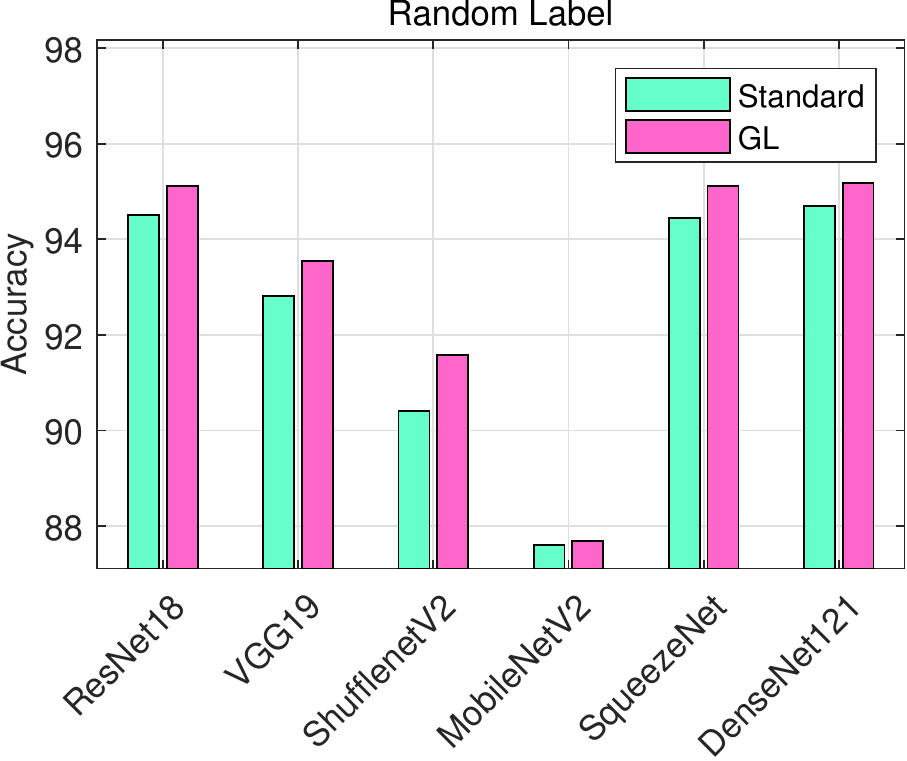}
  \caption{The effect of the network architectures.}\label{fig:net}
\end{figure}

\begin{figure*}
  \centering
  \includegraphics[width=1\textwidth]{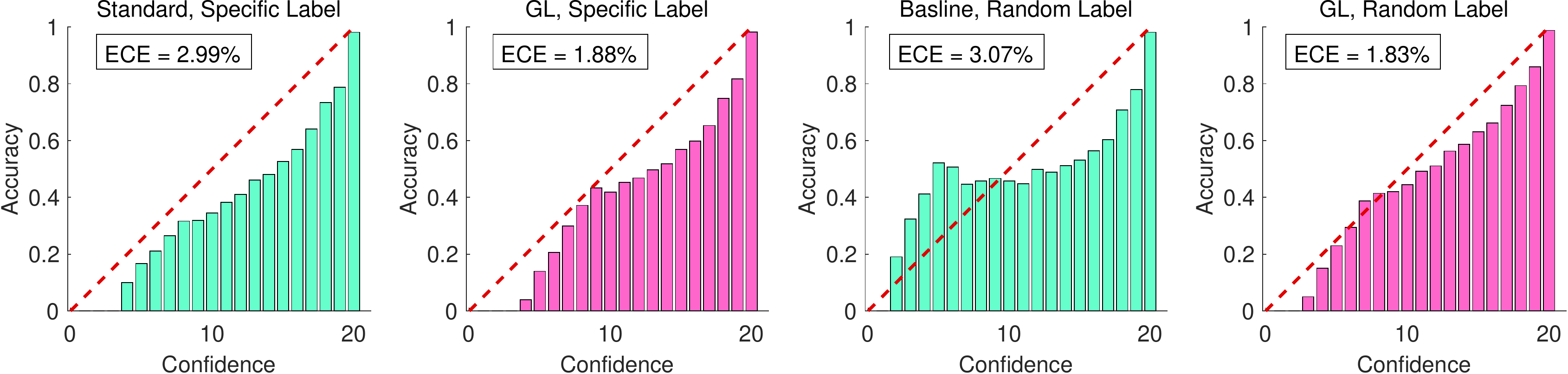}
  \caption{The calibration results on CIFAR10. The confidence is equally divided into 20 intervals, and each bar represents the expected accuracy of samples whose confidence values are in the same interval. The red dotted diagonal indicates the perfect calibration.}
  \label{fig:ece}
\end{figure*}

\subsection{Effect of Out-of-distribution Samples}
We investigate the impact of both the proportion parameter $\alpha$ and the source variations of out-of-distribution samples within the training set. For this analysis, we utilize the CIFAR10 dataset and employ a ResNet18 architecture for training the networks. To assess the efficacy of labeling strategies for out-of-distribution samples, we separately examine the cases of specific and random labels. Our comparative study pits the performance of GL against that of the standard method, which employs a cross-entropy loss function for handling non-IID samples.

\subsubsection{Effect of the proportion $\alpha$}
To explore the effects of a high value of the proportion parameter $\alpha$, we designate samples from SVHN as out-of-distribution samples for the CIFAR10 dataset and vary $\alpha$ within the range of $[0.05, \ldots, 0.5]$. The resulting experimental findings are presented in \figurename~\ref{fig:ood}. For the standard approach, we observe a decline in classification performance as the proportion of out-of-distribution samples increases, regardless of whether specific or random labels are employed. This sensitivity to the presence of out-of-distribution samples confirms the theoretical insights provided in Theorem~\ref{tm:rb}. Specifically, a high value of $\alpha$ exacerbates the divergence between the optimal solution achieved on clean, in-distribution samples and the empirical minimizer obtained using the standard method on non-IID samples. In contrast, the performance of GL remains stable across varying proportions of out-of-distribution samples, indicating its robustness to such variations. During training, the upper bound of $\lambda$, as described in Theorem~\ref{tm:ob}, is progressively reduced to fulfill the criteria specified in Theorem~\ref{tm:gegl}. By adaptively adjusting $\lambda$, GL effectively harmonizes the two terms on the right-hand side of the bound outlined in Theorem~\ref{tm:gegl}. Hence, GL is adept at robustly learning network models from non-IID data sets that include out-of-distribution samples.

\subsubsection{Effect of sources}
We utilize samples from Mini-Imagenet as the out-of-distribution data for the CIFAR10 dataset and partition Mini-Imagenet into ten distinct subsets based on class order, with each subset comprising samples from ten classes. Each of these subsets serves as a unique source of out-of-distribution samples mixed into the training data, and we evaluate the impact of each source individually. The findings are presented in \figurename~\ref{fig:data}. Our results indicate that the choice of out-of-distribution source can significantly influence classification outcomes. This variability can be attributed to the distributional discrepancy $d_{\mathcal{H}}(\mathcal{P}_I, \mathcal{P}_O)$ between in-distribution and out-of-distribution samples, as delineated in Theorem~\ref{tm:rb} and Theorem~\ref{tm:gegl}. Across all the different sources, GL consistently outperforms the standard method, highlighting its robustness to a wide array of out-of-distribution samples. This superior performance is achieved by adaptive adjustment of the upper bound of $r(\theta, x, y)$ for each sample, effectively narrowing the gap as outlined in Theorem~\ref{tm:gegl}.

\subsection{Effect of Network Architectures}
We employ various deep network architectures, including ResNet18, VGG19, ShufflenetV2, MobileNetV2, Senet18, and DenseNet121, to train models using clean data, consisting of in-distribution samples from CIFAR10 and out-of-distribution samples from the first subset of Mini-Imagenet. With a fixed value of $\alpha = 0.1$, we compare the classification performance of GL with the standard method. The findings are summarized in \figurename~\ref{fig:net}. Our results demonstrate that the choice of network architecture influences classification performance, with residual networks (ResNet18 and DenseNet121) showing superior outcomes. Importantly, GL consistently outperforms the standard method across all architectures, achieving improvements of $0.55 \%$ with specific labels and $0.64\%$ with random labels. This attests to the general applicability of GL. The method extends the conventional cross-entropy loss function to accommodate the influence of out-of-distribution samples, making it versatile across different network architectures. Furthermore, as indicated by Theorem~\ref{tm:gegl}, the effectiveness of GL is architecture-agnostic, provided the networks are sufficiently powerful to meet the low upper bound $\lambda$ condition for each training sample.

\begin{figure}
  \centering
  \includegraphics[width=0.24\textwidth]{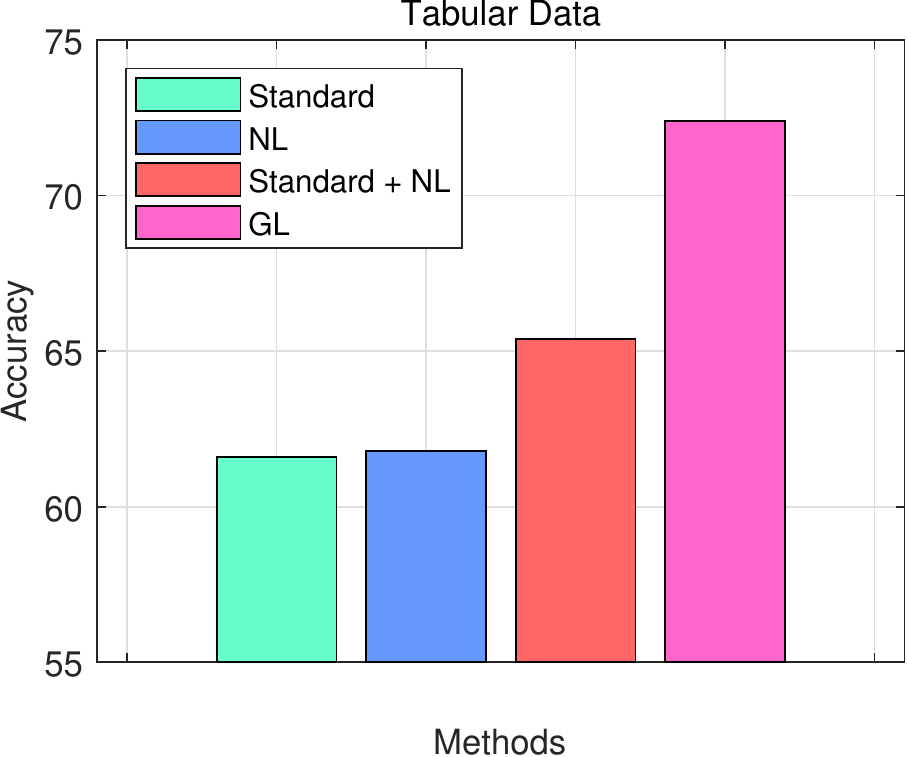}
  \includegraphics[width=0.24\textwidth]{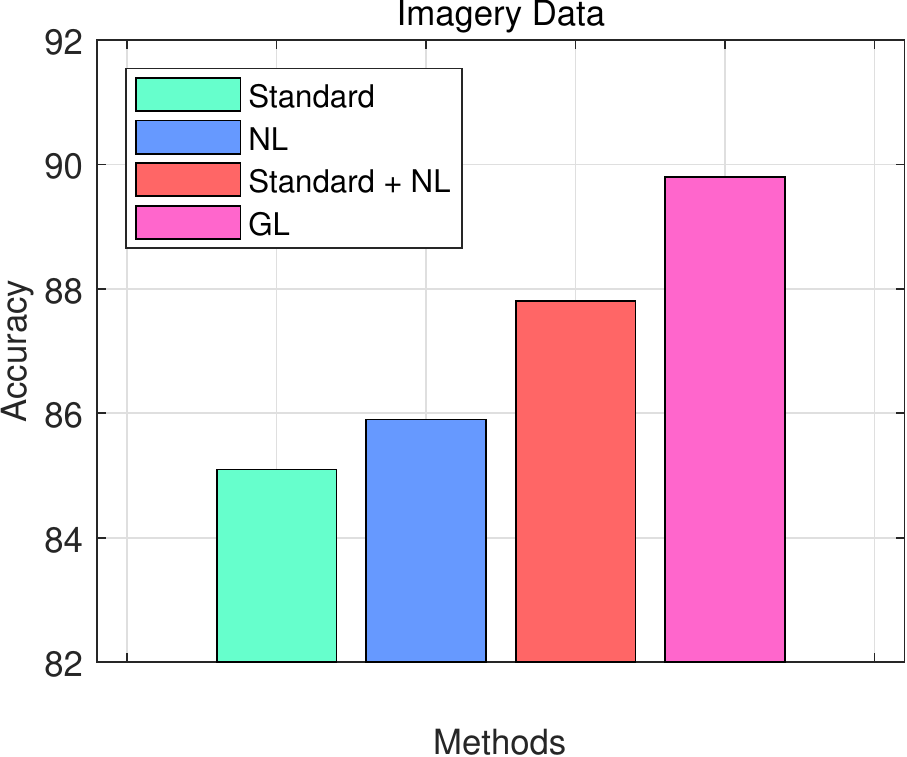}
  \caption{Results of the ablation study. Each bar for tabular data presents the average classification accuracy across the eight tabular datasets. Each bar for imagery data presents the average classification accuracy across the three imagery datasets with specific and random out-of-distribution labels.}
  \label{fig:abla}
\end{figure}

\subsection{Calibration}
We evaluate the ECE of both the standard method and GL on CIFAR10, utilizing the ResNet18 architecture, the first subset of Mini-Imagenet as the out-of-distribution samples, and a component parameter of $\alpha = 0.1$. The findings are depicted in \figurename~\ref{fig:ece}. The standard method registers ECE values of $2.99\%$ and $3.07\%$ for out-of-distribution samples with specific and random labels, respectively. In contrast, GL demonstrates significantly improved calibration with ECE values of $1.88\%$ and $1.93\%$. These results highlight the poor calibration of the standard method and the near-perfect calibration achieved by GL, thereby affirming the reliability of predictions. The standard method is disproportionately impacted by out-of-distribution samples during training, often leading it to produce high-confidence but incorrect predictions. In stark contrast, the robustness of GL to such samples is achieved by judiciously leveraging complementary labels, thereby mitigating the risk of propagating incorrect label information.

\subsection{Ablation Study}
In order to scrutinize the importance of adaptively adjusting the weights for ground-truth and complementary label losses based on prediction confidence, we conduct an ablation study. For context, GL can be conceptualized as dynamically weighting the cross-entropy loss of the standard method and the alternative cross-entropy loss from NL based on prediction confidence. To this end, we compare GL with three other configurations: the standard method, NL, and a composite approach that uniformly blends the losses of standard and NL methods (referred to as Standard + NL). The outcomes of the ablation study across both tabular and imagery datasets are illustrated in \figurename~\ref{fig:abla}.

On tabular data, GL outperforms the standard, NL, and Standard + NL methods by margins of $17.5 \%$, $17.1 \%$ and $10.7 \%$, respectively. Similarly, on imagery data, the respective performance gains for GL are
$5.52 \%$, $4.54 \%$ and $2.28 \%$. These consistent findings across both data types offer valuable insights. When compared to the standard and NL methods, it becomes evident that relying solely on either ground-truth or complementary labels is insufficient for robustly training networks on non-IID data that include both in- and out-of-distribution samples. These methods lack the finesse to differentiate between the two types of samples, implicitly trusting all labels equally. In comparison to the Standard + NL approach, the results suggest that adaptive weighting strategy of GL, which takes prediction confidence into account, is instrumental in its superior performance. This dynamic approach enables GL to focus predominantly on ground-truth labels for in-distribution samples and complementary labels for out-of-distribution samples, thereby widening the confidence gap between these two categories.

\section{Conclusion} \label{sec:con}
This study represents the first effort to train neural networks robustly on non-IID data comprising both in-distribution and out-of-distribution samples. Notably, these two types of samples are intermixed, and the out-of-distribution samples are incorrectly annotated with in-distribution labels. We introduce a novel Gray Learning (GL) approach that adaptively learns from both ground-truth labels and complementary labels, dynamically adjusting the loss weights for each based on prediction confidence. To substantiate the efficacy of GL, we derive generalization bounds rooted in the Rademacher complexity of the hypothesis class, demonstrating that GL yields tighter bounds compared to conventional methods that rely solely on cross-entropy loss. Empirical evaluations confirm that our approach is robust against varying in-distribution and out-of-distribution sample proportions, different sources, and alternative labeling schemes for out-of-distribution samples, outperforming competing methods based on robust statistics. Theoretical results indicate that the discrepancy between in-distribution and out-of-distribution samples is crucial for the generalization bound. Consequently, quantifying and minimizing this discrepancy present intriguing and promising avenues for future research.

% use section* for acknowledgment
%\ifCLASSOPTIONcompsoc
%  \section*{Acknowledgments}
%\else
%  \section*{Acknowledgment}
%\fi

%

\ifCLASSOPTIONcaptionsoff
  \newpage
\fi

\appendices
\section{Proof of Theorem~\ref{tm:rb}}
To proceed, we introduce the following generalization bounds:
\begin{lemma}[\cite{ML:14}]
Assume that for all $x \in \mathcal{X}$, $y \in \mathcal{Y}$ and $h \in \mathcal{H}$, we have that $\vert \mathcal{L}(h(x), y)\vert \leq c$. Let the optimal $h^* = \arg \min_{h \in \mathcal{H}} \epsilon_{\mathcal{P}}(\mathcal{L}, h)$ and $\widehat{h} = \arg \min_{h \in \mathcal{H}} \widehat{\epsilon}_{\mathcal{P}}(\mathcal{L}, h)$. With the probability of at least $1 - \delta$, we have
\begin{align*}
\vert \widehat{\epsilon}_{\mathcal{P}}(\mathcal{L}, h) - \epsilon_{\mathcal{P}}(\mathcal{L}, h) \vert \leq 2 \mathcal{R}(\mathcal{L} \circ \mathcal{H} \circ \mathcal{D}) + 4c \sqrt{\frac{2 \ln(4 / \delta)}{N}}
\end{align*}
where $\mathcal{R}(l \circ \mathcal{H} \circ \mathcal{D})$ is the Rademacher complexity of $\mathcal{H}$ with respect to $\mathcal{L}$ and $\mathcal{D}$.
\label{lm:rc}
\end{lemma}
For any $h \in \mathcal{H}$ and any $\mathcal{L}$
\begin{equation}
\begin{aligned}
& \vert \epsilon_{\mathcal{P}_M}(\mathcal{L}, h) - \epsilon_{\mathcal{P}_I}(\mathcal{L}, h) \vert \\
= & \vert \alpha \epsilon_{\mathcal{P}_O}(\mathcal{L}, h) + (1 - \alpha) \epsilon_{\mathcal{P}_I}(\mathcal{L}, h) - \epsilon_{\mathcal{P}_I}(\mathcal{L}, h) \vert \\
= &  \alpha \vert \epsilon_{\mathcal{P}_O}(\mathcal{L}, h) - \epsilon_{\mathcal{P}_I}(\mathcal{L}, h) \vert \\
\leq & \alpha d_{\mathcal{H}}(\mathcal{P}_I, \mathcal{P}_O).\\
\label{eq:p11}
\end{aligned}
\end{equation}
According to Lemma~\ref{lm:rc}, with the probability of at least $1 - \delta$, for any $h \in \mathcal{H}$,
\begin{equation}
\begin{aligned}
& \vert \epsilon_{\mathcal{P}_M}(\mathcal{L},h) - \widehat{\epsilon}_{\mathcal{P}_M}(\mathcal{L},h) \vert \\
\leq & (1 - \alpha) \vert \epsilon_{\mathcal{P}_I}(\mathcal{L},h) - \widehat{\epsilon}_{\mathcal{P}_I}(\mathcal{L},h) \vert + \alpha \vert \epsilon_{\mathcal{P}_O}(\mathcal{L},h) - \widehat{\epsilon}_{\mathcal{P}_O}(\mathcal{L},h)  \vert \\
\leq & \underbrace{(1 - \alpha) \left( 2 \mathcal{R}(\mathcal{L} \circ \mathcal{H} \circ \mathcal{D}_I) + 4c \sqrt{\frac{2 \ln(8 / \delta)}{N_I}} \right)}_{\triangleq \mathfrak{B}_1(8 / \delta)} \\
& +  \underbrace{\alpha \left( 2\mathcal{R}(\mathcal{L} \circ \mathcal{H} \circ \mathcal{D}_O) + 4c \sqrt{\frac{2 \ln(8 / \delta)}{N_O}} \right)}_{\triangleq \mathfrak{B}_2(8 / \delta)}.
\label{eq:p12}
\end{aligned}
\end{equation}
Applying the bound Eq.~(\ref{eq:p11}) and Eq.~(\ref{eq:p12}), we have the following, with the probability at least $1 - \delta$,
\begin{equation}
\begin{aligned}
& \epsilon_{\mathcal{P}_I}(\mathcal{L},\widehat{h}_M) \\
\leq & \epsilon_{\mathcal{P}_M}(\mathcal{L},\widehat{h}_M) + \alpha d_{\mathcal{H}}(\mathcal{D}_I, \mathcal{D}_O) \\
\leq & \widehat{\epsilon}_{\mathcal{P}_M}(\widehat{h}_M) + \alpha d_{\mathcal{H}}(\mathcal{D}_I, \mathcal{D}_O) + \mathfrak{B}_1(8 / \delta) + \mathfrak{B}_2(8 / \delta)\\
\leq & \widehat{\epsilon}_{\mathcal{P}_M}(h_I^*) + \alpha d_{\mathcal{H}}(\mathcal{D}_I, \mathcal{D}_O) + \mathfrak{B}_1(8 / \delta) + \mathfrak{B}_2(8 / \delta)\\
\end{aligned}
\end{equation}
Now applying the bound Eq.~(\ref{eq:p12}) and Eq.~(\ref{eq:p11}), we have the following with the probability at least $1 - \delta$,
\begin{equation}
\begin{aligned}
& \epsilon_{\mathcal{P}_I}(\mathcal{L},\widehat{h}_M) \\
\leq & \epsilon_{\mathcal{P}_M}(h_I^*) + \alpha d_{\mathcal{H}}(\mathcal{D}_I, \mathcal{D}_O) + 2\mathfrak{B}_1(16 / \delta) + 2\mathfrak{B}_2(16 / \delta)\\
\leq & \epsilon_{\mathcal{P}_I}(h_I^*) + 2 \alpha d_{\mathcal{H}}(\mathcal{D}_I, \mathcal{D}_O) + 2\mathfrak{B}_1(16 / \delta) + 2\mathfrak{B}_2(16 / \delta).\\
\label{eq:p13}
\end{aligned}
\end{equation}
Because the loss function $\mathcal{L}$ is $L$-Lipschitz continuous w.r.t. $h \in \mathcal{H}$. According to the Talagrand's contraction lemma~\cite{FML:18}, we have
\begin{equation}
\begin{aligned}
& \mathcal{R}(\mathcal{L} \circ \mathcal{H} \circ \mathcal{D}_I) \leq L \mathbb{E}_{\sigma \in \{\pm 1\}^{N_I}} \left[ \sup_{h \in \mathcal{H}} \sum_{i = 1}^{N_I} \sigma_i h(x_i^I) \right] \\
& \mathcal{R}(\mathcal{L} \circ \mathcal{H} \circ \mathcal{D}_O) \leq L \mathbb{E}_{\sigma \in \{\pm 1\}^{N_O}} \left[ \sup_{h \in \mathcal{H}} \sum_{j = 1}^{N_O} \sigma_j h(x_j^O) \right]. \\
\label{eq:p14}
\end{aligned}
\end{equation}
To bound $\mathcal{R}(\mathcal{L} \circ \mathcal{H} \circ \mathcal{D}_I)$ and $\mathcal{R}(\mathcal{L} \circ \mathcal{H} \circ \mathcal{D}_O)$ further, we require the following lemma,
\begin{lemma}[\cite{SIC:18}]
Let $\mathcal{H}$ be the class of real-valued networks of depth $d$ over the domain $\mathcal{X}$ and $x \in \mathcal{X}$ is upper bounded by $B$, i.e., for any $x$, $\Vert x \Vert \leq B$. Assume the Frobenius norm of the weight matrices $W_1,\ldots,W_d$ are at most $M_1,\ldots,M_d$. Let the activation function be 1-Lipschitz, positive-homogeneous, and applied element-wise (such as the ReLU). Then,
\begin{equation*}
\mathbb{E}_{\sigma \in \{\pm 1\}^N} \left[ \sup_{h \in \mathcal{H}} \sum_{i = 1}^N \sigma_i h(x_i) \right]\leq \sqrt{N} B(\sqrt{2d \ln2} + 1) \prod_{i = 1}^d M_i.
\end{equation*}
\label{lm:rb}
\end{lemma}
We complete the proof by substituting Eq.~(\ref{eq:p14}) into Eq.~(\ref{eq:p13}) and applying Lemma~\ref{lm:rb}.

\section{Proof of Theorem~\ref{tm:ob}}
Rewriting $\mathcal{L}_{M}(x,y)$, we have
\begin{equation}
\begin{aligned}
\mathcal{L}_{M}(x,y) = & - Q_{\theta}(y|x) \log Q_{\theta}(y|x) \\
& - (1 - Q_{\theta}(y|x)) \sum_{k = 1}^K \log \left( 1 - Q_{\theta}(k|x) \right) \\
& + (1 - Q_{\theta}(y|x)) \log \left( 1 - Q_{\theta}(y|x) \right). \\
\label{eq:tm21}
\end{aligned}
\end{equation}
For the sum of the first and the third terms in Eq.~(\ref{eq:tm21}), we have
\begin{equation}
\begin{aligned}
&(1 - Q_{\theta}(y|x)) \log \left( 1 - Q_{\theta}(y|x) \right)- Q_{\theta}(y|x) \log Q_{\theta}(y|x)\\
= & \log\frac{(1 - Q_{\theta}(y|x))^{1 - Q_{\theta}(y|x)}}{Q_{\theta}(y|x)^{Q_{\theta}(y|x)}}\\
= & \log \frac{\left( Q_{\theta}(y|x) (1 - Q_{\theta}(y|x))\right)^{1 - Q_{\theta}(y|x)}}{Q_{\theta}(y|x)}\\
= & -  \log Q_{\theta}(y|x) + \left( 1 - Q_{\theta}(y|x) \right)\left( Q_{\theta}(y|x) (1 - Q_{\theta}(y|x))\right).
\label{eq:tm22}
\end{aligned}
\end{equation}
We complete the proof by substituting Eq.~(\ref{eq:tm22}) into Eq.~(\ref{eq:tm21}).

\section{Proof of Theorem~\ref{tm:gegl}}
We consider the constraint $r(\theta, x, y) \leq \lambda, \forall (x,y) \sim \mathcal{P}_M$ to bound the Rademacher complexities $\mathcal{R}(\mathcal{L} \circ \mathcal{H} \circ \mathcal{D}_I)$ and $\mathcal{R}(\mathcal{L} \circ \mathcal{H} \circ \mathcal{D}_O)$. Due to the affection of the constraint on exploring the hypothesis class $\mathcal{H}$, we need to obtain the upper bound of $h^y_{\theta}(x), \forall (x,y) \sim \mathcal{P}_M$ to bound the Rademacher complexities.

To process, we calculate the upper bound of $\sum_{k = 1}^K \log \left( 1 - Q_{\theta}(k|x) \right)$ in $r(\theta, x, y)$ with $\sum_{k = 1}^K Q_{\theta}(k|x) = 1$. We solve the constrained optimization problem by forming a Langrangian and introducing Lagrange multiplier $\mu$. Accordingly, we define
\begin{equation}
\begin{aligned}
\mathcal{G} = \sum_{k = 1}^K \log \left( 1 - Q_{\theta}(k|x) \right) + \mu \left( \sum_{k = 1}^K Q_{\theta}(k|x) - 1\right).
\end{aligned}
\end{equation}
The partial derivatives of $\mathcal{G}$ with respect to $Q_{\theta}(k|x)$ and $\lambda$ are
\begin{equation}
\begin{aligned}
\frac{\partial \mathcal{G}}{\partial Q_{\theta}(k|x)} = \frac{1}{Q_{\theta}(k|x) - 1} +  \mu  = 0, \forall k \in [K]\\
\label{eq:tm31}
\end{aligned}
\end{equation}
and
\begin{equation}
\begin{aligned}
\frac{\partial \mathcal{G}}{\partial \lambda} = \sum_{k = 1}^K Q_{\theta}(k|x) - 1 = 0.
\label{eq:tm32}
\end{aligned}
\end{equation}
According to Eq.~(\ref{eq:tm31}) and Eq.~(\ref{eq:tm32}), we can obtain the maximum value of $\mathcal{G}$ when $Q_{\theta}(k|x) = 1 / K, \forall k \in [K]$, i.e.,
\begin{equation}
\begin{aligned}
\sum_{k = 1}^K \log \left( 1 - Q_{\theta}(k|x) \right) \leq K \log \left( 1 - \frac{1}{K} \right).
\label{eq:tm33}
\end{aligned}
\end{equation}
According to the basic inequality $x^y \geq \frac{x}{x + y}, \forall x > 0, y\in(0,1)$ and $Q_{\theta}(y|x) \in (0,1)$, we have
\begin{equation}
\begin{aligned}
& (1 - Q_{\theta}(y|x)) \log Q_{\theta}(y|x) \left( 1 - Q_{\theta}(y|x) \right)\\
= & \log \left(Q_{\theta}(y|x) \left( 1 - Q_{\theta}(y|x) \right)\right)^{(1 - Q_{\theta}(y|x))}\\
\geq & \frac{Q_{\theta}(y|x) \left( 1 - Q_{\theta}(y|x) \right)}{Q_{\theta}(y|x) \left( 1 - Q_{\theta}(y|x) \right) + 1 - Q_{\theta}(y|x)}\\
=& \frac{Q_{\theta}(y|x)}{1 + Q_{\theta}(y|x)} \geq \frac{Q_{\theta}(y|x)}{2}.
\label{eq:tm34}
\end{aligned}
\end{equation}
We obtain the upper bound of $h^y_{\theta}(x)$ by combining Eq.~(\ref{eq:tm33}), Eq.~(\ref{eq:tm34}) and the assumption $\log \sum_{k = 1}^K \exp(h_{\theta}^k(x)) \leq z$,
\begin{equation}
\begin{aligned}
h^y_{\theta}(x) & \leq \log(2\lambda - 2) + z, \forall (x,y) \sim \mathcal{P}_M.
\label{eq:tm35}
\end{aligned}
\end{equation}
According to Jensen's inequality, Khintchine-Kahane inequality~\cite{KKI:12} and Eq.~(\ref{eq:tm35}), we have
\begin{equation}
\begin{aligned}
\mathbb{E}_{\sigma \in \{\pm 1\}^N} \left[ \sup_{h \in \mathcal{H}} \sum_{i = 1}^N \sigma_i h(x_i) \right]
\leq & \sup_{h \in \mathcal{H}} \sqrt{  \sum_{i = 1}^N \Vert h(x_i) \Vert^2}  \\
\leq & LK\sqrt{N}(\log(2\lambda - 2) + z).
\label{eq:tm36}
\end{aligned}
\end{equation}
We complete the proof by substituting Eq.~(\ref{eq:tm22}) and Eq.~(\ref{eq:tm36}) into Eq.~(\ref{eq:tm21}).

\bibliographystyle{IEEEtran}
\bibliography{GL}

\begin{IEEEbiography}[{\includegraphics[width=1in,height=1.25in,clip,keepaspectratio]{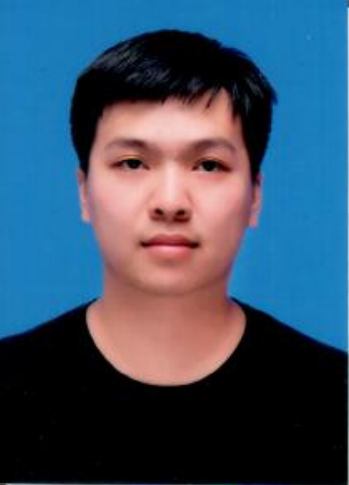}}]{Zhilin Zhao}
received his Ph.D. degree from the University of Technology Sydney in 2022. Before that, he earned his B.S. and M.S. degrees from the School of Data and Computer Science at Sun Yat-Sen University, China, in 2016 and 2018, respectively. He is currently a Post-Doctoral Fellow at Macquarie University, Australia. His research interests encompass generalization analysis, distribution discrepancy estimation, and out-of-distribution detection.

% He has published papers as the first author in international journals including IEEE TPAMI, IEEE TNNLS, and TMLR.
% He has contributed to the field with publications in esteemed international journals such as IEEE TPAMI, IEEE TNNLS, and TMLR.
\end{IEEEbiography}
%\vspace{-5in}

\begin{IEEEbiography}[{\includegraphics[width=1in,height=1.25in,clip,keepaspectratio]{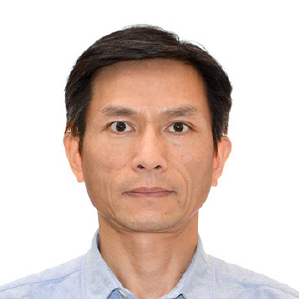}}]{Longbing Cao} %\begin{IEEEbiography}[{\includegraphics[width=1in,height=1.25in,clip,keepaspectratio]{Fig/Cao}}]{Longbing Cao}
received a PhD degree in pattern recognition and intelligent systems at Chinese Academy of Sciences in 2002 and another PhD in computing sciences at University of Technology Sydney in 2005. He is the Distinguished Chair Professor in AI at Macquarie University and an Australian Research Council Future Fellow (professorial level). His research interests include AI and intelligent systems, data science and analytics, machine learning, behavior informatics, and enterprise innovation.
\end{IEEEbiography}
%\vspace{-5in}

\begin{IEEEbiography}[{\includegraphics[width=1in,height=1.25in,clip,keepaspectratio]{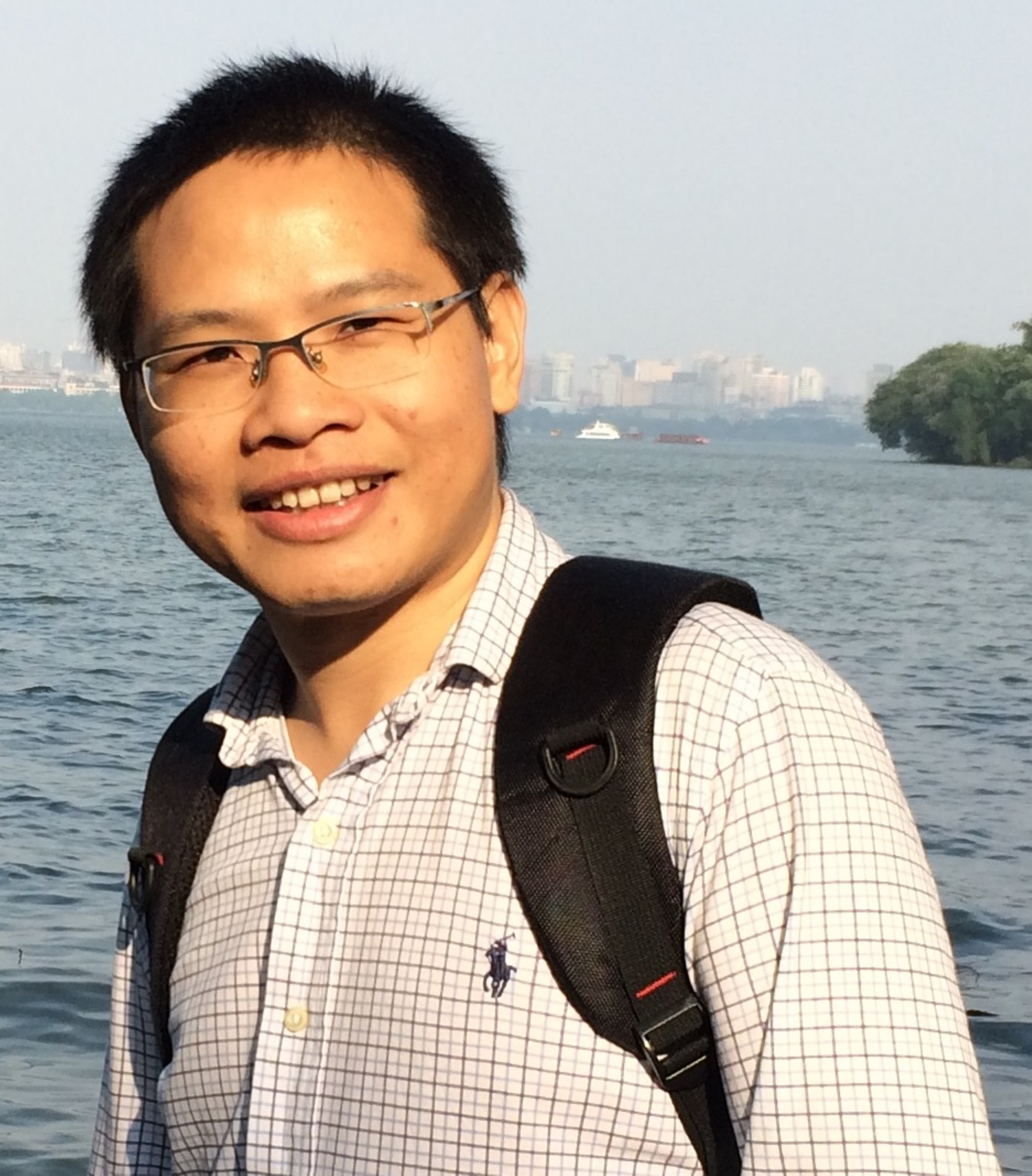}}]{Chang-Dong Wang} received the Ph.D. degree in computer science in 2013 from Sun Yat-sen University, Guangzhou, China. He joined Sun Yat-sen University in 2013, where he is currently an associate professor with School of Computer Science and Engineering. His current research interests include machine learning and data mining. He has published over 80 scientific papers in international journals and conferences such as IEEE TPAMI, IEEE TKDE, IEEE TCYB, IEEE TNNLS, KDD, AAAI, IJCAI, CVPR, ICDM, CIKM and SDM. His ICDM 2010 paper won the Honorable Mention for Best Research Paper Awards. He won 2012 Microsoft Research Fellowship Nomination Award. He was awarded 2015 Chinese Association for Artificial Intelligence (CAAI) Outstanding Dissertation. He is an Associate Editor in Journal of Artificial Intelligence Research (JAIR).
\end{IEEEbiography}

\end{document}